\documentclass{article} 
\usepackage{iclr2024_conference,times}

\usepackage{amsmath,amsfonts,bm}

\def\eqref#1{equation~\ref{#1}}

\def\1{\bm{1}}

\DeclareMathAlphabet{\mathsfit}{\encodingdefault}{\sfdefault}{m}{sl}
\SetMathAlphabet{\mathsfit}{bold}{\encodingdefault}{\sfdefault}{bx}{n}

\usepackage{hyperref}
\usepackage{url}

\usepackage{xcolor}         
\definecolor{sbblue}{HTML}{4878d0}
\definecolor{sbred}{HTML}{d65f5f}
\definecolor{sbgreen}{HTML}{6acc64}
\definecolor{sbbluedeep}{HTML}{4c72b0}
\definecolor{sbreddeep}{HTML}{c44e52}
\definecolor{sbgreendeep}{HTML}{55a868}

\usepackage[utf8]{inputenc} 
\usepackage[T1]{fontenc}    
\usepackage{hyperref}       
\hypersetup{
  colorlinks,
  citecolor=sbblue,
  linkcolor=sbred,
  urlcolor=sbblue}
\usepackage{url}            
\usepackage{booktabs}       
\usepackage{amsfonts}       
\usepackage{nicefrac}       
\usepackage{microtype}      
\usepackage{enumitem}       
\usepackage{siunitx}        
\usepackage{pmboxdraw}      
\usepackage{caption}        

\usepackage{amsthm,xpatch}  

\theoremstyle{remark}

\theoremstyle{definition}

\makeatletter
\newcounter{proofpart}[proof] 

\makeatother                

\usepackage{amsmath}        

\usepackage{graphicx}
\usepackage{xspace}         
\makeatletter
\DeclareRobustCommand\onedot{\futurelet\@let@token\@onedot}
\def\@onedot{\ifx\@let@token.\else.\null\fi\xspace}

\def\eg{e.g\onedot} 
\def\ie{i.e\onedot}

\makeatother

\newcommand{\laionfour}{LAION-400M}
\newcommand{\laiontwo}{LAION-200M}

\newcommand{\imagenet}{ImageNet}
\newcommand{\imagenetk}{ImageNet-1k}
\newcommand{\imagenett}{ImageNet-Train}
\newcommand{\imagenets}{ImageNet-Sketch}
\newcommand{\imagenetv}{ImageNet-Val}

\newcommand{\imageneta}{ImageNet-A}
\newcommand{\imagenetr}{ImageNet-R}
\newcommand{\imagenetvtwo}{ImageNet-V2}

\newcommand{\celeba}{CelebA}
\newcommand{\waterbirds}{Waterbirds}

\newcommand{\resnetfifty}{ResNet-50}
\newcommand{\resnethundred}{ResNet-101}

\newcommand{\vitbthirty}{CLIP~ViT-B/32}
\newcommand{\vitbsixteenp}{CLIP~ViT-B/16+}

\title{Does CLIP’s generalization performance mainly stem from high train-test similarity?}

\author{%
    Prasanna Mayilvahanan$^{1,2,3}$\footnotemark[1] \quad Thaddäus Wiedemer$^{1,2,3}$\footnotemark[1] \quad Evgenia Rusak$^{1,2,3}$\\
    \\
    \textbf{Matthias Bethge}$^{1,2}$\ \quad \textbf{Wieland Brendel}$^{2,3,4}$\\
    \\
    $^1$University of Tübingen \quad $^2$Tübingen AI Center\\
    $^3$Max-Planck-Institute for Intelligent Systems, Tübingen \quad $^4$ELLIS Institute Tübingen\\
    \\
    {\tt\small prasanna.mayilvahanan@uni-tuebingen.de, thaddaeus.wiedemer@gmail.com}
}

\iclrfinalcopy 
\begin{document}
\renewcommand*{\thefootnote}{\fnsymbol{footnote}}
\footnotetext[1]{Equal contribution. Code available at \url{https://github.com/brendel-group/clip-ood}}
\footnotetext[0]{}

\maketitle

\begin{abstract}
Foundation models like CLIP are trained on hundreds of millions of samples and effortlessly generalize to new tasks and inputs. Out of the box, CLIP shows stellar zero-shot and few-shot capabilities on a wide range of out-of-distribution (OOD) benchmarks, which prior works attribute mainly to today's large and comprehensive training dataset (like LAION). However, it is questionable how meaningful CLIP's high zero-shot performance is as it seems likely that web-scale datasets like LAION simply contain many samples that are similar to common OOD benchmarks originally designed for ImageNet. To test this hypothesis, we retrain CLIP on pruned LAION splits that replicate ImageNet’s train-test similarity with respect to common OOD benchmarks. While we observe a performance drop on some benchmarks, surprisingly, CLIP’s overall performance remains high. This shows that high train-test similarity is insufficient to explain CLIP’s performance, and other properties of the training data must drive CLIP to learn good representations. Additionally, by pruning data points that are dissimilar to the OOD benchmarks, we uncover a 100M split of LAION (¼ of its original size) on which CLIP can be trained to match its original performance.

\end{abstract}
\section{Introduction}\label{sec:intro}

Large models like GPT-4~\citep{openai2023gpt4, ChatGPT}, CLIP~\citep{radford2021learning}, or LLaMa~\citep{touvron2023llama} are changing the technological and academic landscape with their unprecedented performance and breadth of viable applications. A core characteristic of these \emph{Foundation Models}~\citep{bommasani2021opportunities} is that they are trained on hundreds of millions or even billions of data points scraped from the internet. For example, OpenCLIP~\citep{schuhmann2022laionb}, the open-source version of CLIP~\citep{radford2021learning}, is trained on \laionfour{}, a web-scale dataset with a wide variety of image-text pairs~\citep{schuhmann2021laion}. CLIP forms the backbone of generative models like DALL-E2~\citep{ramesh2022hierarchical} and is known for its remarkable zero-shot and few-shot performance on a wide range of tasks, specifically on out-of-distribution (OOD) benchmarks like \imagenets{}~\citep{wang2019learning}, \imagenetr{}~\citep{hendrycks2020many}, etc.

Prior work has shown that CLIP's stellar performance stems mainly from its data distribution~\citep{fang2022data, radford2021learning}.
Nevertheless, it remains unclear which specific properties of the training distribution, such as its scale, diversity, density, or relation to the test set, drive performance.
OOD benchmarks like \imagenets{} and \imagenetr{} were initially designed in reference to \imagenetk{}~\citep{imagenet_cvpr09}, which had served as the primary dataset driving progress in machine vision for several years before the emergence of web-scale datasets.
\imagenets{}, \imagenetr{}, and others are considered OOD because they share the same content (\ie, classes) as \imagenetk{} but are \emph{dissimilar} in terms of style, pose, scale, background, or viewpoint. There is no guarantee that these datasets are also \emph{dissimilar} to \laionfour{}. We provide evidence in Fig.~\ref{fig:motivation} where we choose samples from \imagenets{} and \imagenetr{} and examine their nearest perceptual neighbors in \laionfour{} and \imagenett{}. We find highly \emph{similar} neighbors and even exact duplicates in \laionfour{} while neighbors in \imagenett{} are relatively \emph{dissimilar}. In other words, models trained on \laionfour{} may perform well on conventional OOD benchmarks simply due to being trained on semantically and stylistically \emph{similar} data points.
Naturally, the question arises:
\begin{center}
    \emph{Does CLIP's accuracy on OOD benchmarks mainly stem from highly similar images in its train set?}
\end{center}

By \textit{highly similar images}, we mean images that are stylistically and semantically more similar to the test sets than any image in \imagenetk{} is.
To answer this question, we make the following contributions:

\begin{itemize}[leftmargin=0.3cm]
    \item In Sec.~\ref{sec:sim_metric}, we begin by introducing \emph{perceptual similarity}~\citep{ilharco_gabriel_2021_5143773}, which has previously been shown to capture stylistic and semantic similarity between images~\citep{dreamsim,datacomp,tipadapter}.
    We show in Sec.~\ref{sec:nn_sim} that the similarity of nearest neighbors under this metric generally impacts CLIP's performance.
    Specifically, we (i) observe a high correlation between zero-shot accuracy and nearest-neighbor similarity of test samples and (ii) demonstrate that similarity-based pruning of the training set greatly affects CLIP's performance.
    \item Based on these insights, we compare the distribution of nearest-neighbor similarities of different training sets in Sec.~\ref{sec:nn_sim_distribution} and find that they differ substantially. We hypothesize that CLIP's high performance might be largely explained by the training samples that cause this difference, which we term \emph{highly similar images}.
    %
    %
    \item Sec.~\ref{sec:gap} formalizes the notion of \emph{highly similar images} based on the \emph{similarity gap} of two training distributions. Under this formalization, \emph{highly similar images} of \laionfour{} lie within the similarity gap of \imagenett{} to a given test set, \ie, are more similar to test samples than any image in \imagenett{} is. We go on to show how pruning can align the similarity gap of both distributions, such that test sets are as dissimilar to pruned \laionfour{}-splits as they are to \imagenett{}.
    %
    \item As our central result in Sec.~\ref{sec:experiments}, we surprisingly find that training CLIP on the curated subsets only marginally decreases performance on the corresponding OOD benchmarks (Tab.~\ref{tab:main_experiment}). We conclude that high train-test similarity cannot fully explain CLIP's remarkable performance, and other properties of \laionfour{} must play a role.
    \item To facilitate future research into the impact of training on the performance of vision-language foundation models, we curate a 100M subset of \laionfour{} (¼ of its original size) on which CLIP maintains its full OOD benchmark performance (Sec.~\ref{sec:nn_sim} \& ~\ref{sec:appendix:coreset}).
\end{itemize}
\section{Related Work}\label{sec:related}

\paragraph{Measuring OOD generalization} To assess expected model performance in the wild, researchers use different test sets that are considered OOD with respect to the training distribution.
The terms OOD generalization, (distributional) robustness, or just generalization are used interchangeably by the community. This work mainly focuses on standard datasets that share classes with \imagenet{}. They include: image renditions (\imagenetr{}; \citealp{hendrycks2020many}), unusual camera views and object positions (ObjectNet; \citealp{barbu2019objectnet}), images selected to be difficult for \imagenet{}-trained \resnetfifty{}s (\imageneta{}; \citealp{hendrycks2021natural}) and sketches of ImageNet classes (\imagenets{};~\citealp{wang2019learning}). We also consider two datasets commonly considered in-distribution, namely \imagenetv{} \citep{imagenet_cvpr09}, and \imagenetvtwo{} ~\citep{recht2019imagenet}.

\paragraph{ID vs. OOD generalization}
While researchers treat the test sets listed above as OOD with respect to the training distribution when they study robustness, this core assumption is rarely scrutinized. Large-scale language-image models such as CLIP~\citep{radford2021learning}, ALIGN~\citep{jia2021scaling}, or BASIC~\citep{pham2021combined} claim exceptional OOD generalization and zero-shot capabilities. \citet{fang2022data} probe which aspects of the models---like language supervision, cost function, or training distribution---are related to a model's effective OOD robustness and find that differences in the distribution play a key role.
Further, \citet{nguyen2022quality} find that combining data from multiple sources for training interpolates the model's effective robustness on an OOD test set between the performance of the model trained on either data source.
Here, we aim to extend the findings of \citet{fang2022data} and \citet{nguyen2022quality} by evaluating whether high similarity between training and test set is the main driver of CLIP's claimed performance, or whether CLIP is truly better at generalizing across larger distribution shifts. 

{
\noindent
\centering
\includegraphics[width=0.90\textwidth]{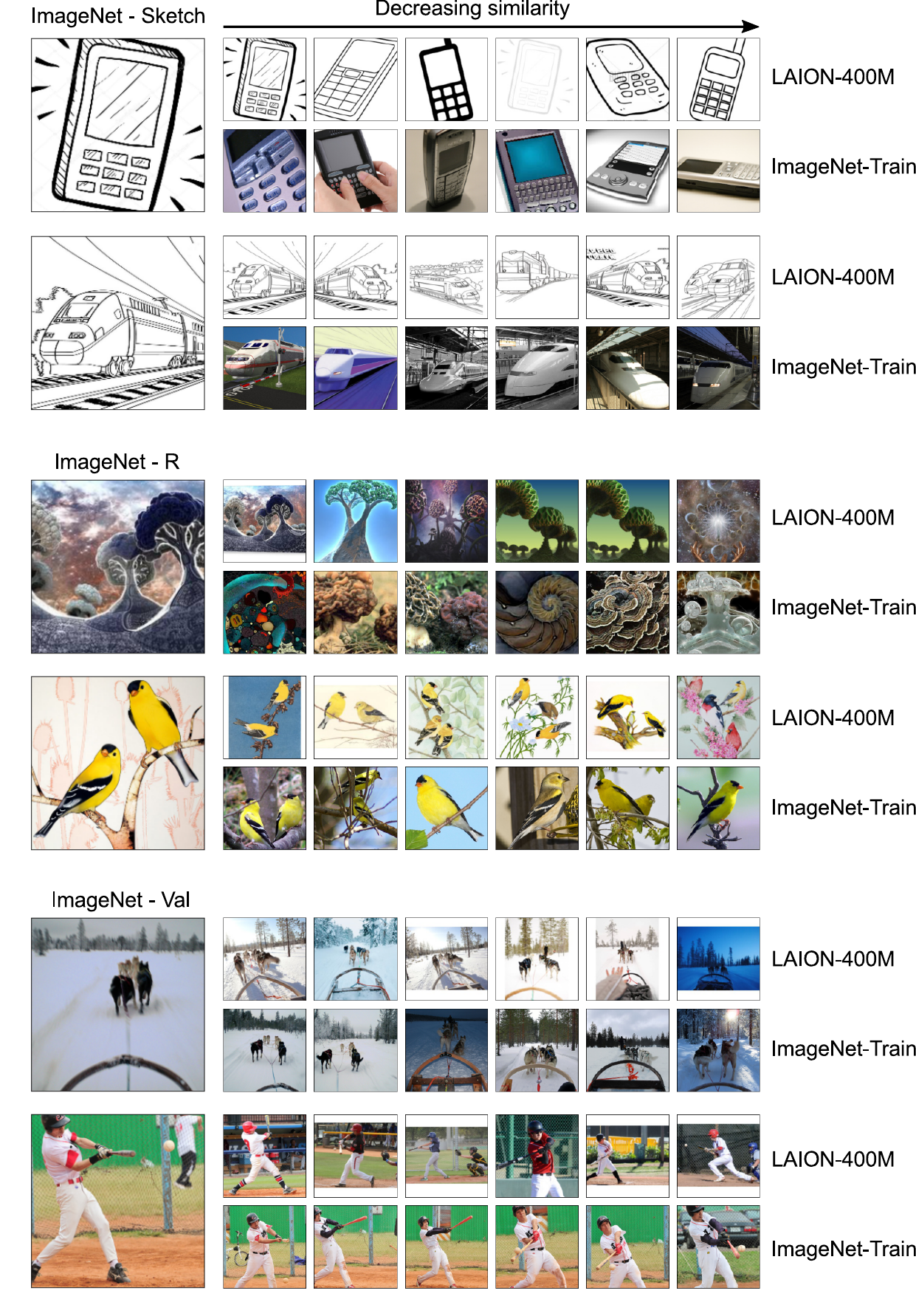}
\vspace{-4mm}
\captionof{figure}{
    \textbf{Similarity of common benchmarks to \laionfour{} and \imagenett{}.}
    We show nearest neighbors of \imagenets{}, \imagenetr{} and \imagenetv{} samples in \laionfour{} and \imagenett{} ordered by decreasing \emph{perceptual similarity}. We omit duplicates within these nearest neighbors. Perceptual similarity is cosine similarity computed in CLIP's image embedding space (see Sec.~\ref{sec:difficulty}) and can be thought of as measuring the perceptual closeness of images in terms of content and style. \laionfour{} clearly contains more similar images to samples from \imagenets{} and \imagenetr{}, in contrast \imagenett{} is more similar to \imagenetv{}. More details in App.~\ref{sec:appendix-nn_vis}.}\label{fig:motivation}
}
\section{Experimental details}\label{sec:exp_details}
This section contains technical specifics of image-to-image similarity computation, training details, deduplication, and \laiontwo{}. Readers can skip this section and return to it when they seek details on the aforementioned. For computing image-to-image similarity, measuring duplicates, and pruning data points, we use \vitbsixteenp{}'s image embedding space. For all our pruning experiments, we train \vitbthirty{}~\citep{dosovitskiy2020image} for 32 epochs with a batch size of 33,600 on one node with eight A100 GPUs (training takes several days, depending on the dataset size). We use the implementation provided by \citet{ilharco_gabriel_2021_5143773} and stick to their settings for learning rate, weight decay, etc. Our downloaded version of \laionfour{} contains only 377M images overall due to missing or broken links, compared to the original 400M used in OpenCLIP~\citep{ilharco_gabriel_2021_5143773}.

\paragraph{\laiontwo{}}
\citet{abbas2023semdedup} show that pruning exact duplicates, near duplicates, and semantically very similar samples \textit{within} \laionfour{} (not yet taking any test sets into account) can reduce dataset size by up to \SI{50}{\%} without performance degradation. We re-implement their method to generate our baseline LAION split containing 199M samples, which we refer to as \laiontwo{}. This step is important to make training multiple instances of CLIP feasible, and we observe that the incurred drop in performance is negligible (compare Tab.~\ref{tab:main_experiment}).
\section{The similarity hypothesis}\label{sec:difficulty}
This section first illustrates how perceptual similarity can be quantified (Sec.~\ref{sec:sim_metric}).
Based on this metric, we demonstrate that CLIP's performance on a test set is strongly related to the \emph{nearest-neighbor similarity} between \laionfour{} and a test set (Sec.~\ref{sec:nn_sim}).
Further, we show that nearest-neighbor similarities differ between \laionfour{} and \imagenett, which leads to the hypothesis that this difference explains CLIP's high classification accuracy on \imagenet-based test sets (Sec.~\ref{sec:nn_sim_distribution}).
Finally, we phrase this hypothesis in terms of \emph{highly similar images}, which leaves us with an interventional method to test this hypothesis (Sec.~\ref{sec:gap}).

\subsection{Quantifying perceptual similarity}\label{sec:sim_metric}
\begin{figure}[t]
    \centering
    \includegraphics[width=\textwidth]{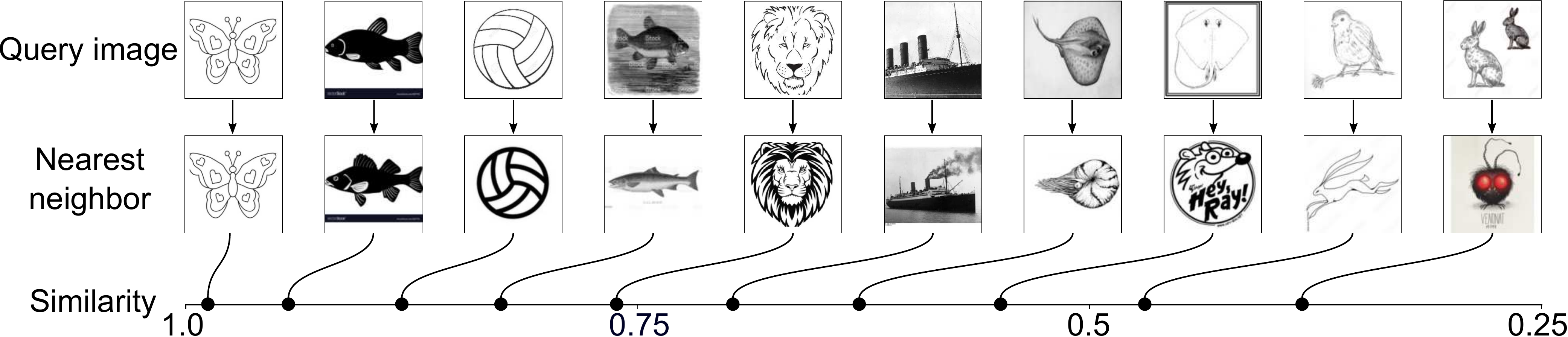}
    \caption{
        \textbf{Relation between \textit{perceptual similarity} and visual closeness of nearest neighbors}. Query images are sampled from \imagenets{} (top row) and are connected to their nearest neighbor in \laionfour{} (bottom row). As in Fig.~\ref{fig:motivation}, perceptual similarity is simply the cosine similarity measured in \vitbsixteenp's image embedding space.}
    \label{fig:sim_metric}
\end{figure}

\citet{abbas2023semdedup} demonstrated that nearest neighbors in the image embedding space of CLIP share \textit{semantic} and \textit{stylistic} characteristics.
We illustrate this in Fig.~\ref{fig:sim_metric}, where we plot samples from \imagenets{} and their nearest neighbors in \laionfour{} for different similarity values.
Visually, the similarity scores correlate well with the closeness of the image pairs.
This is corroborated by other works that demonstrate high perceptual alignment between CLIP's embedding similarity and human perception \citep{dreamsim}, or using it to sample \imagenet{}-like images from a large dataset \citep{datacomp}, or building a similarity-based classifier \citep{tipadapter}.

We follow these works and quantify \emph{perceptual similarity} as the cosine similarity in \vitbsixteenp's image embedding space. App.~\ref{app:comparing_metrics} ablates the choice of the model used to compute this metric. We denote the similarity of two samples $x_i, x_j \in \mathbb R^n$ as
\begin{equation}\label{eq:sim}
    s(x_i, x_j): \mathbb R^n \times \mathbb R^n \to [-1, 1].
\end{equation}
We now consider the relation between a training dataset $\mathcal D$ and a test set $\mathcal T$. Using the similarity metric $s$, we can find the nearest neighbor in the test set for each training sample. This allows us to assign each training sample $x_i \in \mathcal D$ the \emph{nearest-neighbor similarity}
\begin{equation}\label{eq:sim_train}
    s_{\text{train},i}(\mathcal D, \mathcal T) = \max_{t \in \mathcal T} s(t, x_i).
\end{equation}
In the same way, we can assign each test sample $t_i \in \mathcal T$ the \emph{nearest-neighbor similarity}
\begin{equation}\label{eq:sim_test}
    s_{\text{test},i}(\mathcal D, \mathcal T) = \max_{x \in \mathcal D} s(t_i, x).
\end{equation}

\subsection{Nearest-neighbor similarity drives performance}\label{sec:nn_sim}
We can now examine the relationship between nearest-neighbor similarity and CLIP's zero-shot classification performance.

\begin{figure}
    \centering
    \includegraphics[width=\textwidth]{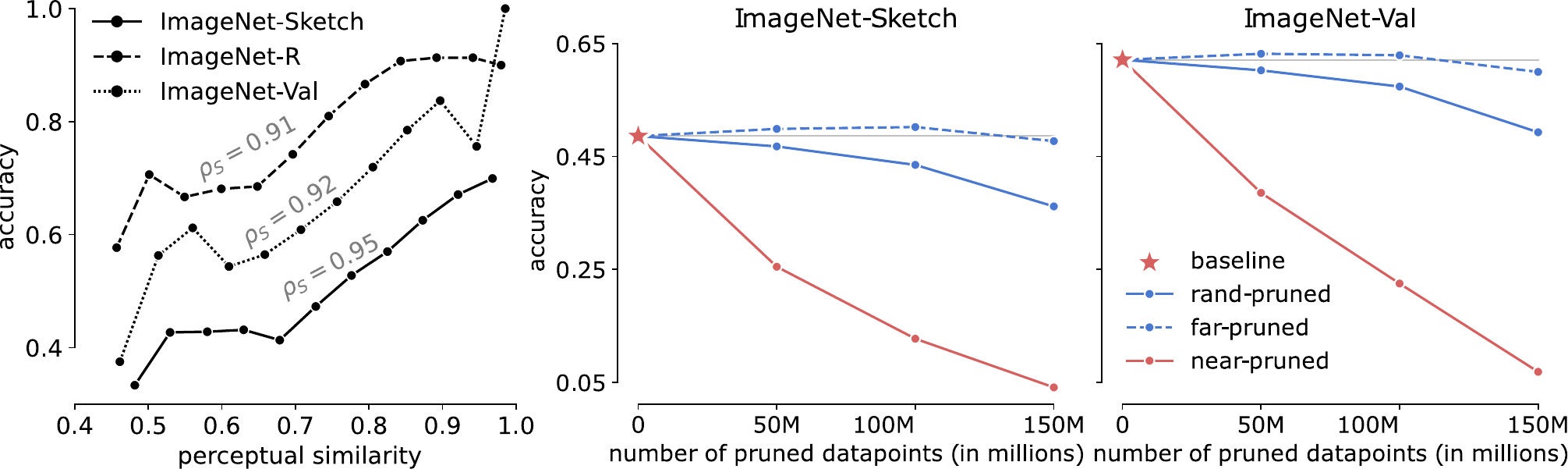}
    \caption{
        \textbf{Nearest-neighbor similarity is predictive of performance}.
        \textbf{Left}: \laionfour{}-trained CLIP's top-1 classification accuracy on test samples is highly correlated to their nearest-neighbor similarity $s_{\text{test},i}$.
        Results are averaged over 0.05 similarity intervals.
        \textbf{Center and right}: Similarity-based pruning greatly impacts CLIP's top-1 classification accuracy.
        We train a baseline model on \laiontwo{} (see Sec.~\ref{sec:exp_details}) and additional models on \laiontwo{}-splits created by random pruning, near-pruning (in order of decreasing similarity), and far-pruning (in order of increasing similarity).
        Compared to training on `rand-pruned' splits (solid blue curve), training on `near-pruned' splits (solid red curve) drastically decreases classification accuracy.
        Training on `far-pruned' splits (dashed blue curve) impacts accuracy comparatively little.
    }\label{fig:correlation_and_pruning}
\end{figure}

Fig.~\ref{fig:correlation_and_pruning}~(left) illustrates that the nearest-neighbor similarity $s_{\text{test},i}$ of test samples in \imagenets{}, \imagenetr{}, and \imagenetv{} to \laiontwo{} is a good predictor of CLIP's top-1 accuracy on these samples. We observe a clear correlation between nearest-neighbor similarity and accuracy across datasets. For \imagenets{}, for example, sketches without similar counterparts in \laionfour{} (similarity 0.38) are classified with \SI{35}{\%} accuracy, while sketches duplicated in \laionfour{}  (similarity close to 1) reach up to \SI{69}{\%} accuracy. We show additional correlation plots for \imagenet-based test sets in App.~\ref{app:additional_experiments} and for other test sets in App.~\ref{app:other_datasets}.

We can observe the impact of nearest-neighbor similarity on classification performance more directly by pruning samples from \laiontwo{} based on their nearest-neighbor similarity $s_{\text{train},i}$ to a given test set, retraining CLIP, and evaluating its zero-shot classification performance on that test set.
We compare three different pruning strategies: `near-pruning' prunes in decreasing order of similarity (pruning samples with high nearest-neighbor similarity first), `far-pruning' prunes in increasing order of similarity, and `rand-pruning' prunes randomly irrespective of similarity.
All strategies produce \laiontwo{}-splits with 50M, 100M, and 150M pruned samples.

CLIP's zero-shot classification performance when trained on these splits is illustrated in Fig.~\ref{fig:correlation_and_pruning} for \imagenets{} and \imagenetv{}.
The `near-pruned' accuracy curve drops much quicker with decreasing dataset size than the `rand-pruned' curve. 
This reiterates that CLIP's classification performance is directly related to the similarity of its training set to the test set.
Additional visualizations for other datasets (both \imagenet{}-based and otherwise) as well as a comparison with \imagenet{}-trained models can be found in Apps.~\ref{app:additional_experiments}~and~\ref{app:other_datasets}.
Note that since we prune large fractions of the training set here, the pruned images are not yet very specific to the test set used to compute $s_{\text{train},i}$. As a result, pruning based on one \imagenet{}-based dataset generally decreases performance across many \imagenet{}-based datasets, although not on not on other tasks (see App.~\ref{app:additional_experiments}).

The observation so far is not surprising: Performance on the test set decreases in tandem with the training distribution's similarity to the test set.
However, our results validate using similarity-based pruning as an effective intervention that allows us to study how training samples impact performance on a given test set.
In the next sections, we will explore how to hone this method to arrive at a more precise conclusion about the role of \emph{highly similar images}.

\paragraph{Core set}
As an aside, we notice that CLIP's performance when trained on `far-pruned' \laiontwo{}-splits remains stable up until a dataset size of 100M (see Fig.~\ref{fig:correlation_and_pruning}).
The performance even slightly surpasses the baseline, further indicating that dissimilar samples do not contribute to CLIP's performance and instead act more like noise in the training data.
Motivated by this performance, we extract a \laionfour{} \emph{core set} with only 100M images by `far-pruning' based on not one but six common \imagenet{}-based benchmarks simultaneously.
CLIP trained on this core set outperforms models trained on a de-duplicated dataset of the same size~\citep{ilharco_gabriel_2021_5143773} and roughly matches the performance of a \laiontwo{}-trained model (see Appx.~\ref{sec:appendix:coreset}).
We release this core set to ease further exploration of the relationship between training distribution and CLIP's zero-shot performance.

\subsection{Comparing nearest-neighbor similarities between training sets}\label{sec:nn_sim_distribution}
\begin{figure}
    \centering
    \includegraphics[width=\textwidth]{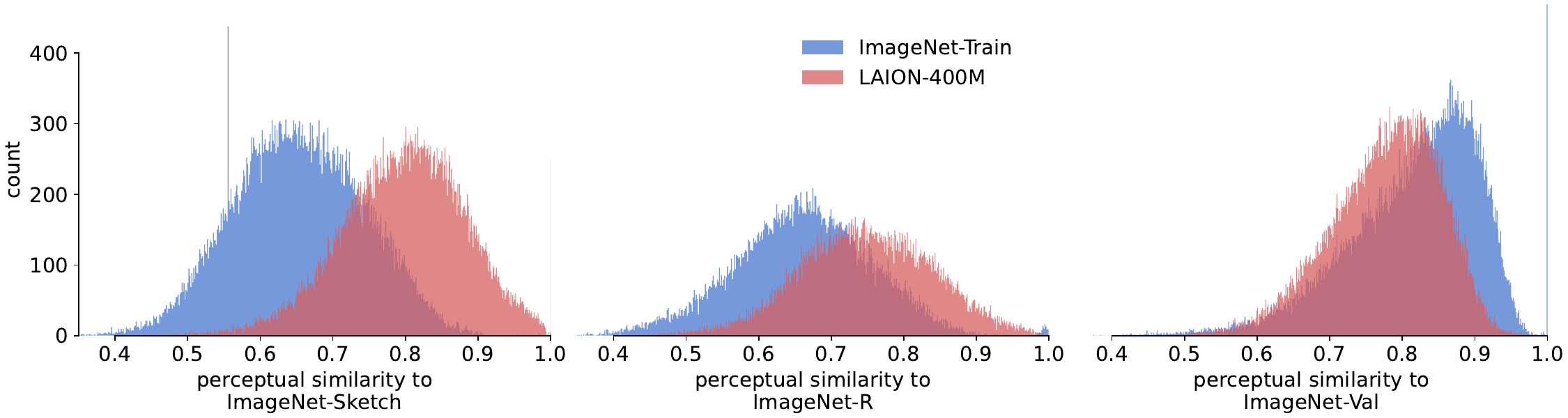}
    \caption{
        \textbf{Nearest-neighbor similarity distribution differs between \laionfour{} and \imagenett{}}.
        The histograms display the similarity $s_{\text{test},i}$ of samples in \imagenets{} (left), \imagenetr{} (center), and \imagenetv{} (right) to their nearest neighbors in \laionfour{} (red) and \imagenett{} (blue). \imagenets{} and \imagenetr{} are overall more similar to \laionfour{}, while \imagenett{} is more similar to \imagenetv{}.
    }\label{fig:histograms}
\end{figure}

Given the impact of nearest-neighbor similarity on CLIP's zero-shot performance, it is natural to ask how \laionfour{}'s nearest-neighbor similarity compares to that of other datasets.
Specifically, for \imagenet{}-based benchmarks like \imagenets{} and \imagenetr{}, we compare the distribution of nearest-neighbor similarities $s_{\text{test},i}$ to \laionfour{} and \imagenett{}.
We have already seen in Fig.~\ref{fig:motivation} that compared to \imagenett{}, \laionfour{} seemed stylistically and semantically much more similar to \imagenets{} and \imagenetr{}, while the effect was reversed for \imagenetv{}.
Using the notion of perceptual nearest-neighbor similarity, we can now fully capture the difference in similarity in a principled manner. This is illustrated in Fig.~\ref{fig:histograms}, where we can now clearly observe that compared to \imagenett{}, \laionfour{} is indeed overall more similar to \imagenets{} and \imagenetr{}.
We show additional histograms for other test sets in Apps.~\ref{app:additional_experiments}~and~\ref{app:other_datasets}.

Moreover, in Appx.~\ref{sec:appendix-duplicates}, we detail how many training samples in \laionfour{} and \imagenett{} are \emph{near duplicates} (duplicates up to small shifts or crops) of the test sets. While we found \SI{3.1}{\%} of \imagenets{} images to have duplicates in \laionfour{}, there are only \SI{0.04}{\%} \imagenets{} duplicates in \imagenett{}. On the other hand, \imagenett{} contains duplicates of \SI{2.67}{\%} \imagenetv{} images as opposed to just \SI{0.14}{\%} \imagenetv{} images  in \laionfour{}.

\laionfour{}-trained CLIP has been reported to outperform \imagenet{}-trained methods on \imagenets{} and \imagenetr{}, while underperforming on \imagenetv{} (see Tab.~\ref{tab:main_experiment}).
In light of the above observation, this could well be explained not by \laionfour{}'s general scale and diversity but specifically by its fraction of training samples whose nearest-neighbor similarity to the test set surpasses that of \emph{any} sample in \imagenett{}.
We term those samples \emph{highly similar images}.
The following section formalizes this concept and explains how we can refine the similarity-based pruning from Sec.~\ref{sec:nn_sim} to quantify their impact on CLIP's zero-shot classification performance.

\subsection{Similarity gap and highly similar images}\label{sec:gap}
Secs.~\ref{sec:nn_sim}~and~\ref{sec:nn_sim_distribution} provide direct and indirect evidence that CLIP's performance on common \imagenet{}-based benchmarks might mainly stem from images in its training set that are \emph{highly similar} to the test sets.
We now formalize this notion and describe how to systematically test our hypothesis.
To this end, we note that even for \imagenett{}, the nearest-neighbor similarity $s_{\text{test},i}$ differs across test samples.
Our goal is to prune \laionfour{} so that the pruned dataset replicates the nearest-neighbor similarities $s_{\text{test},i}$ of \imagenett{}.

\begin{figure}[t]
    \centering
    \includegraphics[width=\textwidth]{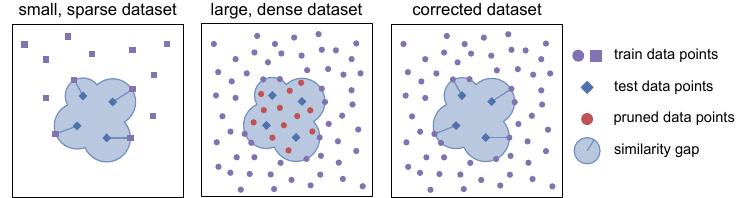}
    \caption{
        \textbf{Aligning the similarity gap of two datasets}.
        A larger, denser, more diverse dataset likely contains samples more similar to given test points than a smaller, sparser one. To control for this, we compute the nearest-neighbor similarity of each test point to the smaller dataset (left) and prune points from the larger dataset that lie within this hull (center). We end up with a corrected large dataset replicating the \textit{similarity gap} of the small one (right).
    }\label{fig:method}
\end{figure}

Let us consider that we now have two training datasets, denoted as $\mathcal{D}_S$ (small, like \imagenett{}) and $\mathcal{D}_L$ (large, like \laionfour{}), and still use a test dataset $\mathcal{T}$ (like \imagenets{}).
For the sake of simplicity, we assume that $\mathcal{D}_S$ is a subset of $\mathcal{D}_L$.
We choose a similarity measure $s$ as in Sec.~\ref{sec:nn_sim}.
We collect all nearest-neighbor similarities $s_{\text{test},i}$ (recall Eq.~\ref{eq:sim_test}) in the set
\begin{equation}
    S(\mathcal D, \mathcal T) = \left\{s_{\text{test},i}(\mathcal D, \mathcal T) \mid i \in \big[1, \left|\mathcal{T}\right| \big] \right\}
\end{equation}
which we term \emph{similarity gap}.
We can think of this set as a full characterization of the training set's similarity to any point in the test set; compare Fig.~\ref{fig:method}.

Based on the assumption that the large dataset contains all samples from the small dataset, it follows that $s_i(\mathcal{D}_S) \leq s_i(\mathcal{D}_L)$.
In other words, the nearest-neighbor similarity to samples in the small training set is always smaller than or equal to the similarity to samples in the large training set.
Consequently, on a per-sample basis, $S(\mathcal D_L, \mathcal T)$ is strictly larger than $S(\mathcal D_S, \mathcal T)$, \ie, the large dataset is generally more similar to the test than the small dataset.
We aim to identify a maximally large subset $\tilde{\mathcal{D}}_L\subseteq\mathcal{D}_L$ of the large training set, such that its similarity gap $S(\mathcal{\tilde D}_L, \mathcal T)$ is equal to the similarity gap $S(\mathcal D_S, \mathcal T)$ of the small dataset (on a per-sample basis, meaning $s_i(\mathcal{\tilde D}_S) = s_i(\mathcal{D}_S)$ for all samples).
To achieve this, we examine each test sample $t_i$ and remove any sample $x\in \mathcal{D}_L$ for which the similarity $s(t_i, x) > s_i(\mathcal{D}_S)$.
We illustrate this procedure in Fig.~\ref{fig:method}.

This method allows us to surgically remove \emph{highly similar images} with respect to a given test set and reference training set.
Compared to the unconstrained pruning in Sec.~\ref{sec:nn_sim}, this will remove far less samples from \laionfour{}, and thus allows us to isolate the impact of \emph{highly similar images}.
\section{Correcting for highly similar images}\label{sec:experiments}
We now apply the framework from Sec.~\ref{sec:gap} to remove highly similar images from \laiontwo{}. 
To ensure that \imagenett{} and \laiontwo{} have the same similarity gap to the test sets, we include all \imagenett{} images in \laiontwo{} with the caption "a photo of a \{object class\}". We refer the reader to Appx. Sec.~\ref{app:choice_imagenet} for a discussion on the choice of ImageNet for our experiments. 

As described in Sec.~\ref{sec:gap}, we first compute the similarity gaps of the smaller dataset, \ie, \imagenett{}, to the samples in each of the six test sets. Pruning \laiontwo{} to these similarity gaps leaves us with six different base splits as shown in Tab.~\ref{tab:main_experiment}. We also generate a `combined-pruned' split that ensures an \imagenett{}-like similarity gap to all test sets simultaneously. We can now train CLIP from scratch on these splits to obtain a corrected zero-shot performance and compare it to the accuracy of CLIP trained by OpenAI and OpenClip~\citep{ilharco_gabriel_2021_5143773, radford2021learning}.

\begin{table}[t]
    \centering
    \caption{\textbf{Corrected zero-shot performance of CLIP ViT-B/32.} `X-pruned' represents a pruned dataset from \laiontwo{} + \imagenet{} such that the similarity gap to 'X' is the same as the similarity gap of \imagenet{} to `X'. The sizes of these subsets are subtracted from the \laiontwo{} + \imagenet{}'s size. Here, `X' is one of the six standard \imagenet{} test sets. `combined-pruned' splits ensure a similarity gap of \laiontwo{} and \imagenett{} to all 6 test sets. CLIP's corrected zero-shot performance drops the most on \imagenets{} and \imagenetr{} with a relative performance drop of \SI{10.8}{\%} and \SI{4.8}{\%} respectively. \textcolor{sbred}{Red} color indicates a drop in performance on the respective test set, and \textcolor{sbblue}{blue} represents a rise. Overall, high performance indicates that highly similar images do not play a key role in explaining CLIP's generalization ability.}\label{tab:main_experiment}
    \vspace{-6pt}
    \sisetup{
        tight-spacing=true,
        detect-family=true,
        detect-weight=true,
        mode=text,
        reset-text-shape=false  
    }
    \setlength{\tabcolsep}{0.4em}
    \renewcommand{\bfseries}{\fontseries{b}\selectfont} 
    \newrobustcmd{\B}{\bfseries}   
    \robustify{\itshape}
    \begin{tabular}{l S[table-format=-8] S[table-format=2.2] S[table-format=2.2] S[table-format=2.2] S[table-format=2.2] S[table-format=2.2] S[table-format=2.2]}
        \toprule
        & & \multicolumn{6}{c}{\textbf{Top-1 Accuracy}} \\
        \cmidrule{3-8}
        \B Dataset & {\B Size} & {\B Val} & {\B Sketch} & {\B A} & {\B R} & {\B V2} & {\B ON} \\
        \midrule
        OpenAI~\citep{radford2021learning} & 400000000 & 63.38 & 42.32 & 31.44 & 69.24 & 55.96 & 44.14\\
        L-400M~\citep{schuhmann2021laion} & 413000000 & 62.94 & 49.39 & 21.64 & 73.48 & 55.14 & 43.94\\
        L-200M     & 199824274 & 62.12 & 48.61 & 21.68 & 72.63 & 54.16 & 44.80 \\
        \midrule
        L-200M + IN-Train & 200966589 & 68.66 & 50.21 & 23.33 & 72.9 & 59.7 & 43.99 \\
        \cmidrule{2-8}
        ├─ val-pruned   & \itshape -377340 & \color{sbred}68.62 & 49.58 & 23.47 & 72.74 & 59.47 & 45.08 \\
        ├─ sketch-pruned   & \itshape -8342783 & 68.34 & \color{sbred}44.78 & 22.7 & 69.35 & 59.52 & 44.12 \\
        ├─ a-pruned   & \itshape -138852 & 68.85 & 50.25 & \color{sbred}22.99 & 72.44 & 60.05 & 44.43 \\
        ├─ r-pruned   & \itshape -5735749 & 68.71 & 46.92 & 23.44 & \color{sbred}69.48 & 59.6 & 45.08 \\
        ├─ v2-pruned   & \itshape -274325 & 68.79 & 50.45 & 23.19 & 72.58 & \color{sbblue}59.84 & 45.33 \\
        ├─ objectnet-pruned  & \itshape -266025 & 68.75 & 50.14 & 22.70 & 72.82 & 59.37 & \color{sbred}43.73 \\
        └─ combined-pruned & \itshape -12352759 & \color{sbred}68.05 & \color{sbred}44.12 & \color{sbred}22.15 & \color{sbred}67.88 & \color{sbred}58.61 & \color{sbblue}44.39 \\
        \bottomrule
    \end{tabular}
\end{table}

The first important point to note in Tab.~\ref{tab:main_experiment} is that for `sketch-pruned' and `r-pruned' datasets, we prune 8.3M and 5.7M samples, respectively. For all other datasets, we prune only around 250K-380K samples. We saw indications of this already in Sec. \ref{sec:difficulty} when we looked at the distribution of nearest-neighbor similarities, see also Tab.~\ref{tab:perc_laion_greater_in}. The number of pruned samples is also highly correlated with the respective accuracies. For CLIP trained on the `r-pruned' dataset and CLIP trained on the `sketch-pruned' dataset, we observe a \SI{4.8}{\%} relative performance decrease on \imagenetr{} and \SI{10.8}{\%} relative performance decrease on \imagenets{} compared to the baseline. There is also a considerable performance change on \imagenetr{} for `sketch-pruned' and on \imagenets{} for `r-pruned'. This is reasonable as there is some style overlap in \imagenets{} and \imagenetr{}. For the other four base splits, we see less than \SI{1}{\%} relative performance change on all six evaluation sets. The performance of the CLIP model trained on the `combined-pruned' split is lower than the baseline on all six eval sets, with sizeable drops in \imagenetr{} and \imagenets{}. We also observe similar trends when we do not add \imagenett{} to the pruned datasets (refer to Tab.~\ref{tab:app_main_exp_wo_imagenet} in the Appx.).
\section{Discussion}\label{sec:discussion}
We now return to our original question: \emph{Does CLIP's accuracy on OOD benchmarks mainly stem from highly similar images in its train set?} To give a definitive answer, we take a closer look at the CLIP model trained on `sketch-pruned'. This model's training set is as dissimilar to \imagenets{} as is \imagenett{}. It features an accuracy of \SI{68.34}{\%} on \imagenetv{}. According to \imagenett{}'s \textit{effective robustness line}~\citep{fang2022data}, at this performance level, we would expect an accuracy of roughly \SI{14}{\%} on \imagenets{}. Instead, we find an accuracy of \SI{44.78}{\%}. In other words, training on a much larger dataset while keeping the similarity gap constant drastically increases generalization performance for CLIP (in this case, by a staggering \num{30} percentage points). This effect is even higher for other datasets. \emph{This indicates that CLIP's impressive performance is not so much the result of a high train-test similarity but that CLIP leverages its dataset scale and diversity to learn more generalizable features.}

\paragraph{What drives generalization?}
Generalization of vision-language models is a complex subject where several factors like architectural choices, caption quality, training procedures, and data distribution play a role. We focus on the training distribution since prior works have studied the effect of the aforementioned factors on CLIP's generalization performance~\citep[\eg,][]{santurkar, mintun2021interaction} and identified it as a prominent factor~\citep{fang2022data}.
Many distribution properties could contribute to generalization performance, but based on raw visualizations of the involved datasets, highly similar images are clearly \emph{a} factor.
Our results only show that it is not the most salient factor and a large chunk of performance remains to be explained. We leave the scrutiny of other likely factors like data diversity and density for future work.
Our work should be interpreted as a step towards finding specific data properties that dictate generalization.

\paragraph{Measuring the true OOD performance}
Our analysis excluded training images from LAION with a smaller similarity gap to test images compared to ImageNet Train. Another interesting analysis would be to prune LAION images to measure its true OOD performance. To remove all images of a certain domain, we need to be able to label each image as `ID' or `OOD'. This essentially means that we need access to a domain classifier (which would also need near-perfect accuracy so that no images are overlooked). Even for the `sketch' domain, where a classifier could conceivably be trained, it is unclear exactly how the classifier should demarcate this domain: Should the domain contain all sketches, even sketches with characteristics not present in ImageNet-Sketch? What about tattoos or small sketches on objects in natural images? For other benchmarks, such as ImageNet-A, it is even less clear how the test images constitute a well-separable domain of images. This vagueness in defining a domain based on a given test set prevents us from building a fair OOD setting, which is why we do not analyze or claim to analyze this. 

\paragraph{Similarity metric}
We defer the reader to Sec.~\ref{app:comparing_metrics} for a discussion and ablation on the choice of \vitbsixteenp{} as the similarity metric.

\paragraph{Highly similar images}
We want to clarify further the notion of \textit{highly similar images}. In Secs.~\ref{sec:sim_metric}, ~\ref{sec:nn_sim}, and~\ref{sec:nn_sim_distribution}, when we use the notion of \textit{similar images} to a given image sample, we refer to images with high perceptual similarity values with no precise constraint. In contrast, in Secs.~\ref{sec:gap}~and~\ref{sec:experiments} we impose a constraint that defines \textit{highly similar images} to a sample as images that are closer to \laiontwo{} than \imagenett{} based on our perceptual similarity metric.

\paragraph{Does compositionality drive performance?}
In this work, we found that high train-test similarity is insufficient to explain CLIP’s high generalization performance on OOD test sets. In our analysis, we only excluded images that were highly similar to the training set to maintain the same similarity gap with respect to ImageNet Train, e.g. sketches of dogs if the test image was a sketch of a dog. However, sketches of other animals and objects still remained in CLIP’s training set. An open question remains whether compositionality~\citep{wiedemer2023compositional} can close the gap between the object and its domain, i.e. whether CLIP can generalize from sketches of cats and natural images of dogs to understanding sketches of dogs.

\section{Conclusion}\label{sec:conclusion} 
CLIP has demonstrated unprecedented performance on common OOD benchmarks designed originally for ImageNet. Given that the training dataset of CLIP is so large and diverse, it is natural to wonder whether its performance stems from the sheer similarity of many training samples to the benchmarks. To the best of our knowledge, we are the first to systematically test if high train-test similarity dictates CLIP’s generalization performance. In our work, we address this by pruning away samples from the training set that are more similar to the test sets than ImageNet samples. Models trained on the pruned dataset do not significantly lose performance and still exhibit stellar generalization capabilities far beyond performance-matched ImageNet-trained models. This indicates that high similarity to the test sets alone can not explain CLIP's generalization ability. We hope this result will prompt the community to investigate other factors that allow models to learn more generalizable features from web-scale datasets.
\section*{Reproducibility Statement}
 For all the basic details of training, pruning, similarity computation, and other analysis, we defer the reader to Sec.~\ref{sec:exp_details}. Details of computing the similarities and its correlation to accuracy is given in the caption of Figs.~\ref{fig:sim_metric},~\ref{fig:correlation_and_pruning}, and Sec.~\ref{sec:sim_metric}. To perform the experiment that observes the effect of 'near-pruning' and 'far-pruning', we defer the reader to Sec.~\ref{sec:nn_sim} and the caption of Fig.~\ref{fig:correlation_and_pruning}. The core methodology of our paper is clearly elucidated in Section~\ref{sec:gap}. Furthermore, the details of generating the datasets and training the models are given in the first and second paragraph of Sec.~\ref{sec:experiments}, and in the caption of Tab.~\ref{tab:main_experiment}. 
\section*{Author contributions}
The project was led and coordinated by PM. The method was jointly developed by PM, TW, with insights from ER, WB, MB. PM conducted all the experiments based on code jointly implemented by PM and TW. PM, TW, ER, and WB jointly wrote the manuscript with additional insights from MB. ER created all figures and visualizations with TW's help using data provided by PM and with comments from WB. 
\section*{Acknowledgments}
We would like to thank (in alphabetical order): Thomas Klein, George Pachitariu, Matthias Tangemann, Vishaal Udandarao, Max Wolff, and Roland Zimmermann for helpful discussions, feedback, and support with setting up the experiments. This work was supported by the German Federal Ministry of Education and Research (BMBF): Tübingen AI Center, FKZ: 01IS18039A. WB acknowledges financial support via an Emmy Noether Grant funded by the German Research Foundation (DFG) under grant no. BR 6382/1-1 and via the Open Philantropy Foundation funded by the Good Ventures Foundation. WB is a member of the Machine Learning Cluster of Excellence, EXC number 2064/1 – Project number 390727645. This research utilized compute resources at the Tübingen Machine Learning Cloud, DFG FKZ INST 37/1057-1 FUGG. We thank the International Max Planck Research School for Intelligent Systems (IMPRS-IS) for supporting PM, TW, and ER.
\bibliography{iclr2024_conference}
\bibliographystyle{iclr2024_conference}

\newpage
\appendix
\section{Distributional dissimilarities of \laiontwo{} and \imagenet{}}
\subsection{nearest-neighbor similarity between LAION / \imagenett{} and other OOD datasets}\label{sec:appendix-nn_sims}

\begin{figure}[h!]
    \centering
     \includegraphics[width=0.75\textwidth]{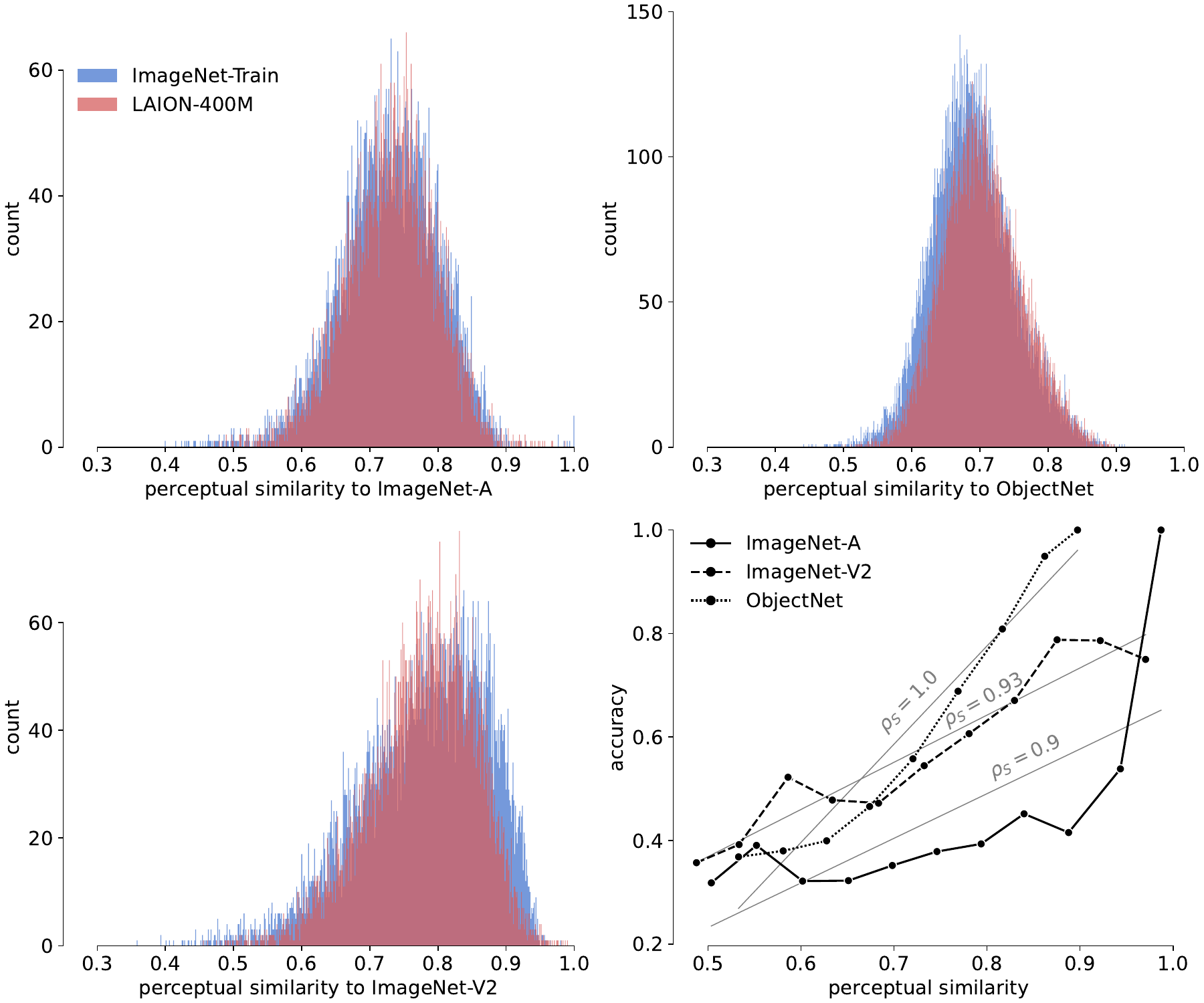}
     \caption{\textbf{Similarity of nearest neighbors to test sets varies between LAION-400M and ImageNet-
Train and is correlated with performance.} Histograms over the nearest-neighbor similarity of test sets \imageneta{} (top left), ObjectNet (top right), and \imagenetvtwo{} (bottom left) to training sets \laionfour{} (red) and \imagenett{} (blue). There are no significant differences in nearest-neighbor similarity for these test sets. Nearest-neighbor similarity of test points to \laionfour{} samples and top-1 classification accuracy is strongly correlated (bottom right). Data points in the correlation plot are averaged over bins (interval = 0.05) of the red histograms.}
     \label{fig:histograms_app}
 \end{figure}

As an extension of our analysis in Sec.~\ref{sec:difficulty}, we plot the nearest-neighbor similarity between \imagenett{}/\laionfour{} and other OOD test sets, namely \imageneta{}~\citep{hendrycks2021natural}, ObjectNet~\citep{barbu2019objectnet} and \imagenetvtwo{}~\citep{recht2019imagenet}, and display our results in Figure~\ref{fig:histograms_app}. There are no significant differences in nearest-neighbor similarity for these test sets. Similar to our results in Figure~\ref{fig:correlation_and_pruning}~(left), we find a strong correlation between perceptual similarity to \laionfour{} and the top-1 accuracy of our LAION-trained model. 

\subsection{Duplicates}\label{sec:appendix-duplicates}
We do a duplicate analysis in Tab.~\ref{tab:duplicates}. To estimate the number of test points with \textit{near duplicates}, we project the test set and \laionfour{} to CLIP's image embedding space and check if any point in \laionfour{} lies in the vicinity ($\epsilon=0.05$) of each of the query test points.

 \begin{table}[h!]
    \centering
    \caption{\textbf{Number of test points of OOD datasets for which we find \textit{near duplicates} in \imagenett{} and \laionfour{}}. A data point is considered \textit{near duplicate (semantic duplicate)} if the distance in the CLIP embedding space is less than 0.05 \citep{abbas2023semdedup}.}\label{tab:duplicates}
    \sisetup{
        tight-spacing=true,
        detect-family=true,
        detect-weight=true,
        mode=text
    }
    \setlength{\tabcolsep}{0.4em}
    \renewcommand{\bfseries}{\fontseries{b}\selectfont} 
    \newrobustcmd{\B}{\bfseries}   
    \begin{tabular}{l l S[table-format=3M] S[table-format=2.2] S[table-format=2.2]}
        \toprule
        & & \multicolumn{2}{c}{\textbf{Duplicates}} \\
        \cmidrule{3-4}
        \B Dataset & \B  Size & {\B ImageNet-Train} & {\B \laionfour{}} \\
        \midrule
        ImageNet-Val  & 50000 & 1336 & 70 \\
        ImageNet-Sketch  & 50889 & 18 & 1553 \\
        ImageNet-R  & 30000 & 104 & 297 \\
        ImageNet-A  & 7500 & 10 & 5 \\
        ImageNet-V2  & 10000 & 10 & 24 \\
        ObjectNet  & 18574 & 0 & 0 \\
        \bottomrule
    \end{tabular}
    \vspace{10pt}
\end{table}

\section{Additional experimental results}\label{app:additional_experiments}
\subsection{Impact of near/far-pruning on all datasets}\label{sec:appendix:near_far}
\begin{figure}[h!]
    \centering
 \includegraphics[width=\textwidth]{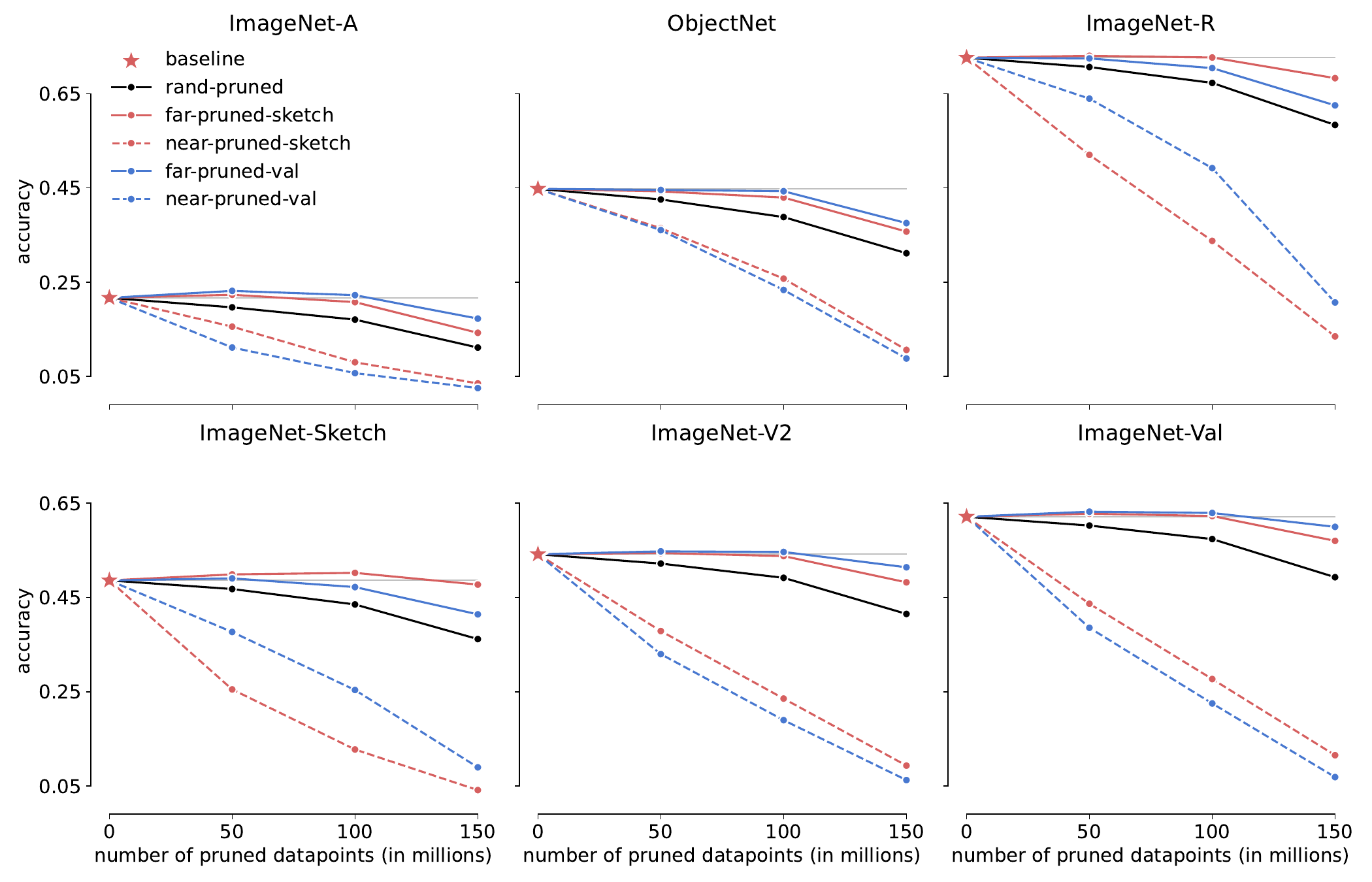}
     \caption{\textbf{The effect of `near-pruning’ and `far-pruning’ with ImageNet-Sketch or ImageNet-Val as the query dataset on the performance of all six test sets.} CLIP’s zero-shot accuracy as a function of the number of pruned points from LAION-200M. The baseline model is trained on de-duplicated LAION-400M, which we call LAION-200M. To generate the ‘near-pruned’ datasets, we remove the images in the decreasing order of similarity (based on CLIP image-embedding similarity) to each of the test sets ImageNet-Sketch and ImageNet-Val, respectively. In contrast, the `far-pruned’ datasets are generated by dropping images in the increasing order of similarity values to the respective test sets. For the `rand-pruned’ datasets, we prune random points.}
     \label{fig:near_far_appendix}
 \end{figure}

In Sec.~\ref{sec:nn_sim}, by using each of the test datasets \imagenetv{} and \imagenets{} and near/far-pruning \laiontwo{}, we trained models and reported the performance on the test datasets, respectively. We now plot the performance of these models on all six datasets in Figure~\ref{fig:near_far_appendix}. `Near-pruning' (`far-pruning') with \imagenets{} results in lower (higher) performance than `near-pruning' (`far-pruning') with \imagenetv{} on \imagenetr{} and \imagenets{}. Likewise, `near-pruning' (`far-pruning') with \imagenetv{} results in lower (higher) performance than `near-pruning' (`far-pruning') with \imagenets{} on \imagenetv{}, \imagenetvtwo{}, and ObjectNet. This is expected because \imagenets{} is characteristically closer to \imagenetr{}, and \imagenetv{} is closer to \imagenetvtwo{}, ObjectNet, and \imageneta{}.

\subsection{Impact of near/far pruning on non-ImageNet-like datasets}
\label{app:near_far_mnist}

\begin{figure}[tb]
    \centering
 \includegraphics[width=\textwidth]{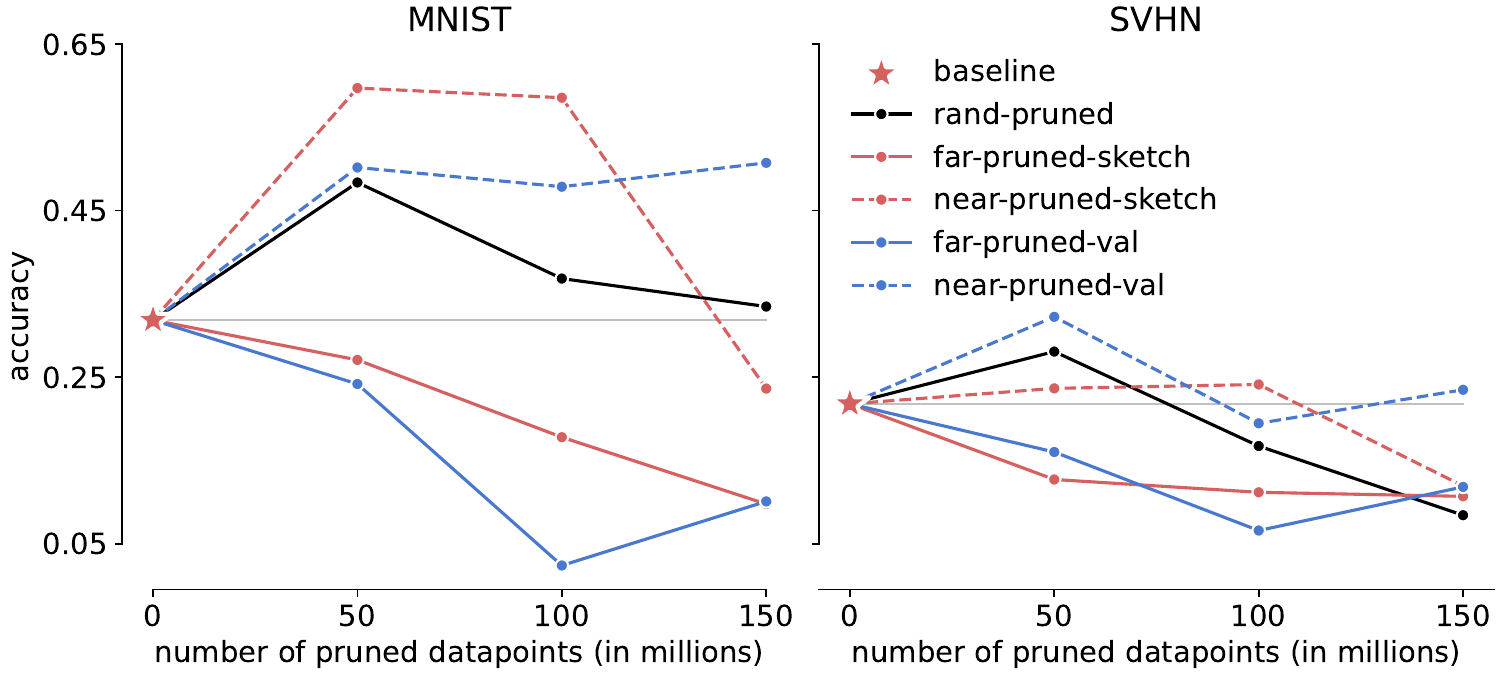}
     \caption{\textbf{The effect of `near-pruning’ and `far-pruning’ with ImageNet-Sketch or ImageNet-Val as the query dataset on the performance on MNIST / SVHN} CLIP’s zero-shot accuracy as a function of the number of pruned points from LAION-200M. The baseline model is trained on de-duplicated LAION-400M, which we call LAION-200M. To generate the ‘near-pruned’ datasets, we remove the images in the decreasing order of similarity (based on CLIP image-embedding similarity) to each of the test sets ImageNet-Sketch and ImageNet-Val, respectively. In contrast, the `far-pruned’ datasets are generated by dropping images in the increasing order of similarity values to the respective test sets. For the `rand-pruned’ datasets, we prune random points.}
     \label{fig:near_far_appendix_mnist}
 \end{figure}

In Fig.~\ref{fig:near_far_appendix}, we observe a consistent trend across all datasets that near-pruning with respect to either \imagenets{} or \imagenetv{} decreases performance while performance is stable (at times even increases) when doing far-pruning.
These findings can be explained by two
hypotheses: 
\begin{enumerate}
    \item The pruned images in the near-pruning setting are reasonably similar to \imagenets{} or \imagenetv{}, thus we see a drop in performance when we train CLIP on the pruned datasets.
    \item Near-pruning with respect to \imagenets{} or \imagenetv{} results in pruning datapoints that are of the highest quality samples from LAION which will perform well on any downstream task, when trained upon.
\end{enumerate}
To decide between these two hypotheses, we repeat the analysis on two test sets which are very dissimilar from ImageNet: SVHN \citep{svhn} and MNIST \citep{deng2012mnist}.
We show the results in Fig.~\ref{fig:near_far_appendix_mnist} and indeed observe a reversed trend compared to Fig.~\ref{fig:near_far_appendix}: Now, for a several cases, near-pruning  increases performance and far-pruning decreases it, respectively.
Thus, the near-pruned datapoints are \textbf{not} comprised of high quality samples which improve performance on all downstream tasks, and pruning away images similar to either \imagenets{} or \imagenetv{} only decreases performance on ImageNet-like datasets.

\subsection{Near/far pruning experiments on \imagenett{}}\label{app:near_far_imagenet}

\begin{figure}[tb]
    \centering
    \includegraphics[width=\textwidth]{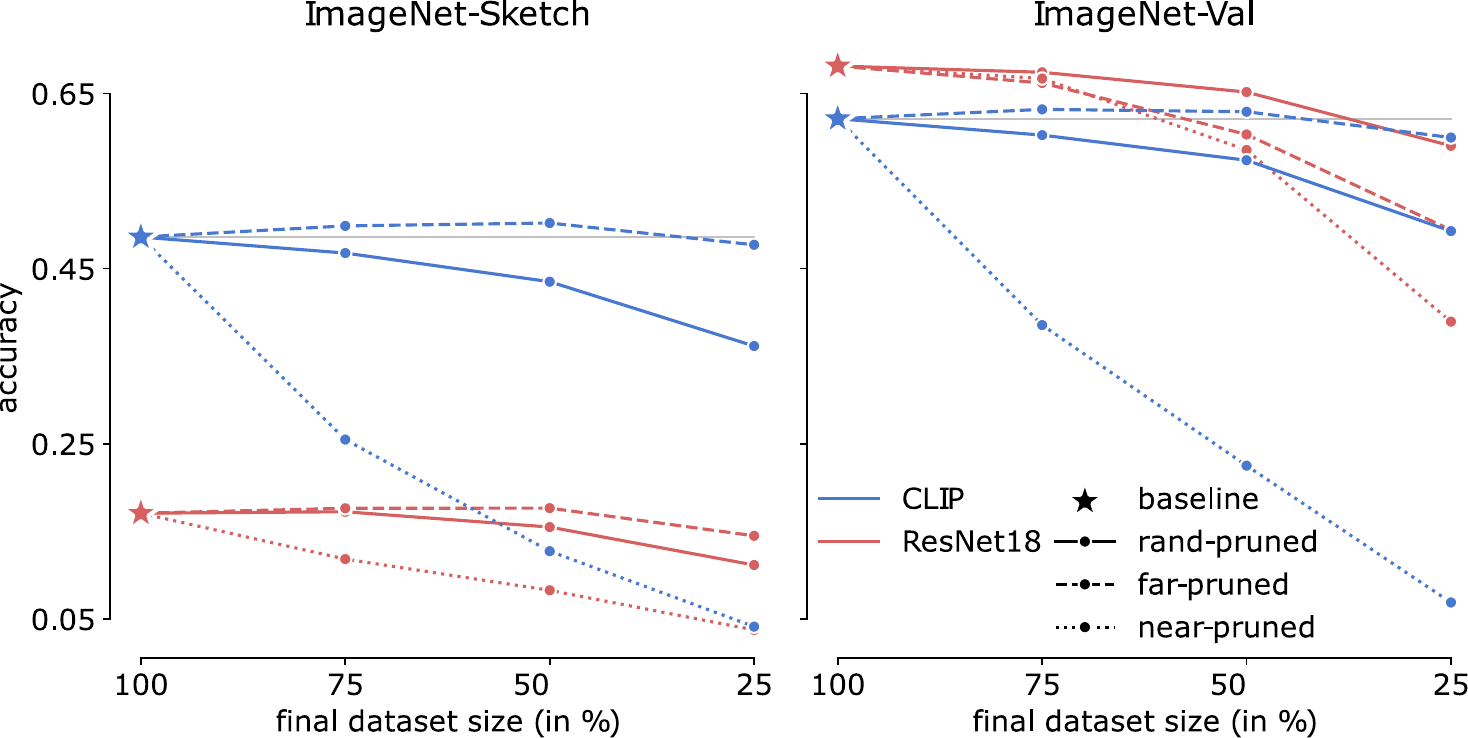}
    \caption{\textbf{Effect of pruning similar and dissimilar points to a given test set on both CLIP and ResNet performance}. The baseline model of CLIP is trained on de-duplicated \laionfour{}, which we call \laiontwo{}. The baseline model of ResNet18 is trained on \imagenett{}. To generate the `near-pruned' datasets, we remove images in decreasing order of similarity to \imagenets{} or \imagenetv{} (based on CLIP image-embedding similarity). In contrast, the `far-pruned' datasets are generated by pruning images in the increasing order of similarity values to the respective test sets. For the `rand-pruned' datasets, we prune random points. Pruning similar images adversely affects performance compared to pruning dissimilar or random images. Generally, near/far pruning affects CLIP more than ResNet18.}
    \label{fig:near_far_comp_abs}
\end{figure}

\begin{figure}[tb]
    \centering
    \includegraphics[width=\textwidth]{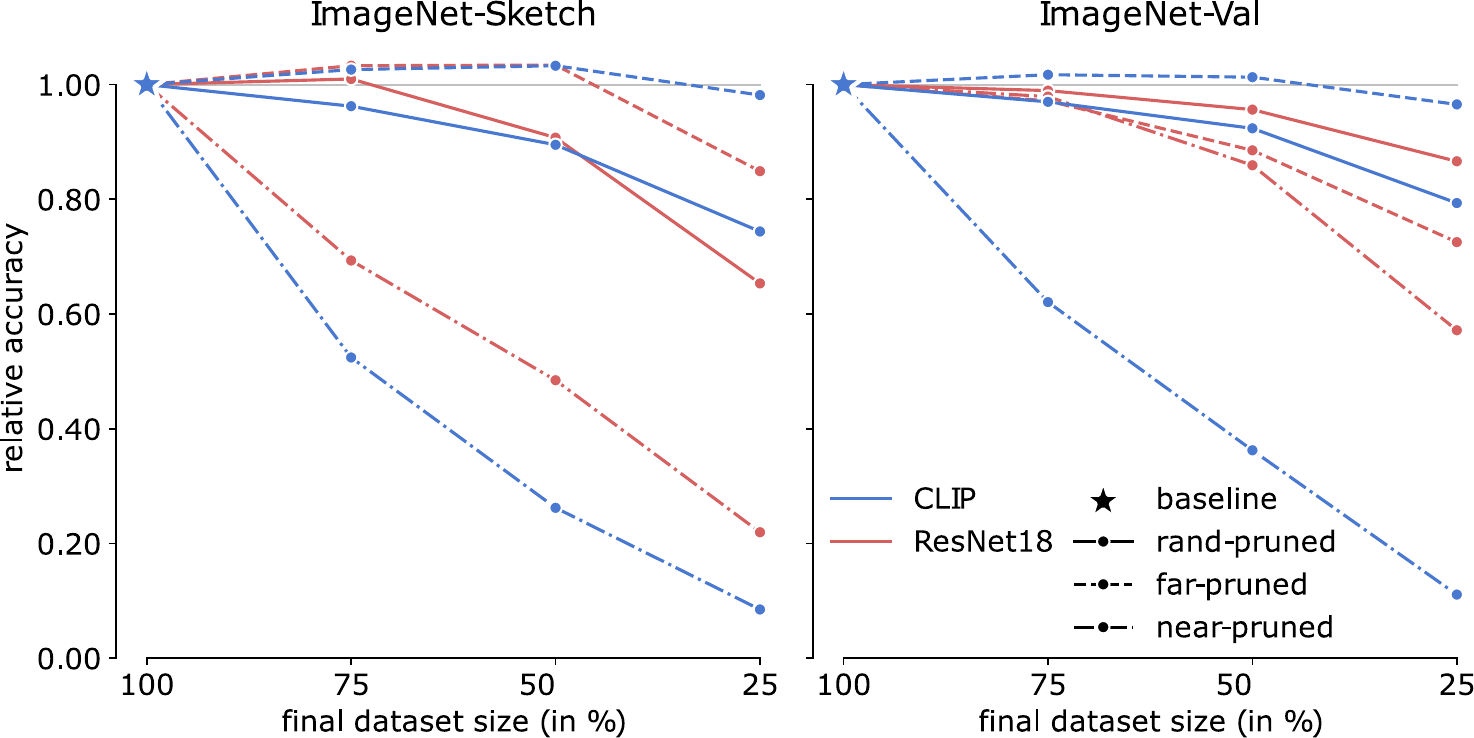}
    \caption{\textbf{Effect of pruning similar and dissimilar points to a given test set on both CLIP and ResNet's performance relative to the baseline}. The baseline model of CLIP is trained on de-duplicated \laionfour{}, which we call \laiontwo{}. The baseline model of ResNet18 is trained on \imagenett{}. To generate the `near-pruned' datasets, we remove images in decreasing order of similarity to \imagenets{} or \imagenetv{} (based on CLIP image-embedding similarity). In contrast, the `far-pruned' datasets are generated by pruning images in the increasing order of similarity values to the respective test sets. For the `rand-pruned' datasets, we prune random points. Pruning similar images adversely affects performance compared to pruning dissimilar or random images. The relative performance drop curves indicate that near/far pruning affects CLIP more than ResNet18.}
    \label{fig:near_far_comp_rel}
\end{figure}

We generate new datasets by near/far pruning datapoints on \imagenett{} with test datasets \imagenets{} and \imagenetv{}. The similarities are computed in the pre-trained \vitbsixteenp{} embedding space. We then train ResNet18s until convergence using standard PyTorch hyperparameter settings. We report the absolute and relative performance (to the respective baseline) in Figure~\ref{fig:near_far_comp_abs} and ~\ref{fig:near_far_comp_rel}. We observe that near/far pruning affects CLIP performance more than ResNet18. 

\subsection{Core set of 100M}\label{sec:appendix:coreset}
In Sec.~\ref{sec:nn_sim}, we identify a 100M core set of \laionfour{}, which, when trained on, leads to a CLIP model that nearly matches the performance of a \laionfour{} trained CLIP model on the six test datasets. Motivated by the performance increase of the `far-pruning' technique in the previous results, we now build several core sets of 100M, which, when trained on, roughly match the performance of CLIP trained on \laionfour{}. Instead of pruning from the farthest point to samples in just a single test set in \vitbsixteenp{}'s embedding space, we now prune from the farthest point to samples from a collection of test sets (all six \imagenetk{} OOD test sets). We do far-pruning with all of the test sets on both \laiontwo{} and \laionfour{} to obtain datasets that we call `all-far-pruned'. For comparison, we also add the performance of CLIP trained on far-pruned datasets with query datasets as \imagenets{} and \imagenetv{}, which we call `sketch-far-pruned' and `val-far-pruned', respectively.

We report the results in Tab.~\ref{tab:far_pruning} and observe that models trained on all of the splits are within \SI{3}{\%} average accuracy range of CLIP trained on \laionfour{}. The model with the highest average accuracy is trained on `all-far-pruned (L-200M)', which is a dataset generated by pruning far or dissimilar images in \laiontwo{} with all 6 test datasets as query datasets. This model also performs better than a model trained on a dataset of the same size generated by the pruning technique SemDeDup \citep{abbas2023semdedup}. SemDeDup aims to prune semantically similar data with minor loss in test performance. We do not suggest this coreset as an alternative to other pruning or deduplication methods that are largely agnostic to the downstream test datasets. Instead, we here created a coreset that is specifically designed to perform well on six OOD test sets to facilitate further research into what aspects drive generalization.
\begin{table}[t]
    \centering
    \caption{\textbf{Performance of `far-pruned' CLIP (ViT-B/32) on the six test sets.} We do `far-pruning' on \laionfour{} with all 6 test sets as query sets and obtain the dataset `all-far-pruned (L-400M).' Similarly, we do `far-pruning' on \laionfour{} with all with all 6 test sets as query sets, \imagenets{}, and \imagenetv{} to get the datasets `all-far-pruned (L-200M)', `sketch-far-pruned (L-200M)', and `val-far-pruned (L-200M)' respectively. These models are compared to off the shelf CLIP model \citep{ilharco_gabriel_2021_5143773}, model trained on \laiontwo{}, and a model trained on SemDeDup \citep{abbas2023semdedup} dataset of size 100M.}\label{tab:far_pruning}
    \sisetup{
        tight-spacing=true,
        detect-family=true,
        detect-weight=true,
        mode=text
    }
    \setlength{\tabcolsep}{0.4em}
    \renewcommand{\bfseries}{\fontseries{b}\selectfont} 
    \newrobustcmd{\B}{\bfseries}   
    \begin{tabular}{l l l l l l S[table-format=3M] S[table-format=2.2] S[table-format=2.2]}
        \toprule
        & & \multicolumn{6}{c}{\textbf{Top-1 Accuracy}} \\
        \cmidrule{3-9}
        \B Dataset & {\B Size} & {\B Val} & {\B Sketch} & {\B A} & {\B R} & {\B V2} & {\B ON} & {\B Avg.} \\
        \midrule
        L-400M & 400M & 62.94 & 49.39 & 21.64 & 73.48 & 55.14 & 43.94 & \textbf{\textit{51.09}}\\
        L-200M & 199.8M & 62.12 & 48.61 & 21.68 & 72.63 & 54.16 & 44.80 & \textit{50.67} \\
        \midrule
        all-far-pruned (L-400M) & 100M & 61.90 & 48.11 & 19.43 & 70.14 & 53.11 & 39.30 & \textit{48.67} \\
        all-far-pruned (L-200M) & 100M & 62.80 & 49.23 & 21.6 & 72.3 & 54.72 & 43.64 & \textit{\textbf{50.71}} \\
        val-far-pruned (L-200M) & 100M & 62.79 & 47.53 & 21.65 & 70.40 & 54.35 & 43.70 & \textit{50.07} \\
        sketch-far-pruned (L-200M) & 100M & 62.27 & 50.21 & 20.77 & 72.67 & 53.77 & 42.95 & \textit{50.44} \\
        \midrule
        SemDeDup  &  100M & 52.19 & 41.70 & 16.71 & 67.05 & 44.96 & 39.59 & \textit{43.7} \\
        \bottomrule
        \end{tabular}
    \vspace{10pt}
\end{table}

\subsection{Main experiments without adding ImageNet-Train}
We repeat the experiments in Sec.~\ref{sec:experiments} without adding \imagenett{} to \laiontwo{} and report results in Tab.~\ref{tab:app_main_exp_wo_imagenet}. We observe the same trends as in Tab.~\ref{tab:main_experiment}.
\begin{table}[t]
    \centering
    \caption{\textbf{Corrected zero-shot performance of CLIP ViT-B/32.} `X-pruned' represents a pruned dataset from \laiontwo{} such that the similarity gap to `X' is roughly the same as the similarity gap of \imagenet{} to `X'. The sizes of these subsets are subtracted from the \laiontwo{}'s size. Here, `X' is one of the six standard \imagenet{} test sets. `combined-pruned' splits ensure a similarity gap of \laiontwo{} and \imagenett{} to all 6 test sets. CLIP's corrected zero-shot performance drops the most on \imagenets{} and \imagenetr{} with a relative performance drop of \SI{11.08}{\%} and \SI{5.99}{\%} respectively. \textcolor{sbred}{Red} color indicates a drop in performance on the respective test set. Overall, high performance indicates that highly similar images do not play a key role in explaining CLIP's generalization ability.}\label{tab:app_main_exp_wo_imagenet}
    \sisetup{
        tight-spacing=true,
        detect-family=true,
        detect-weight=true,
        mode=text,
        reset-text-shape=false  
    }
    \setlength{\tabcolsep}{0.4em}
    \renewcommand{\bfseries}{\fontseries{b}\selectfont} 
    \newrobustcmd{\B}{\bfseries}   
    \robustify{\itshape}
    \resizebox{\textwidth}{!}{
        \begin{tabular}{l l S[table-format=-8] S[table-format=2.2] S[table-format=2.2] S[table-format=2.2] S[table-format=2.2] S[table-format=2.2] S[table-format=2.2]}
            \toprule
            & & & \multicolumn{6}{c}{\textbf{Top-1 Accuracy}} \\
            \cmidrule{4-9}
            \B Model & \B Dataset & {\B Size} & {\B Val} & {\B Sketch} & {\B A} & {\B R} & {\B V2} & {\B ObjectNet} \\
            \midrule
            ViT-B/32  & OpenAI & 400000000 & 63.38 & 42.32 & 31.44 & 69.24 & 55.96 & 44.14\\
            ViT-B/32   & L-400M & 413000000 & 62.94 & 49.39 & 21.64 & 73.48 & 55.14 & 43.94\\
            \midrule
            ViT-B/32   & L-200M              &          199824274 & 62.12 & 48.61 & 21.68 & 72.63 & 54.16 & 44.80 \\
            \cmidrule{3-9}
            ViT-B/32   & ├─ val-pruned       & \itshape   -377340 & \color{sbred}62.12 & 48.38 & 21.45 & 72.2 & 54.76 & 42.79 \\
            ViT-B/32   & ├─ sketch-pruned    & \itshape  -8342783 & 61.55 & \color{sbred}43.22 & 22.28 & 69.6 & 53.53 & 42.77 \\
            ViT-B/32   & ├─ a-pruned         & \itshape   -138852 & 62.49 & 48.49 & \color{sbred}21.63 & 72.15 & 54.38 & 43.25 \\
            ViT-B/32   & ├─ r-pruned         & \itshape  -5735749 & 61.73 & 45.66 & 21.67 & \color{sbred}68.28 & 54.1 & 42.90 \\
            ViT-B/32   & ├─ v2-pruned        & \itshape   -274325 & 62.48 & 48.62 & 22.13 & 72.3 & \color{sbred}53.83 & 43.38 \\
            ViT-B/32   & ├─ objectnet-pruned & \itshape   -266025 & 62.30 & 49.03 & 22.64 & 72.90 & 54.21 & \color{sbred}42.80 \\
            ViT-B/32   & └─ combined-pruned  & \itshape -12352759 & \color{sbred}61.5 & \color{sbred}41.97 & \color{sbred}21.72 & \color{sbred}67.25 & \color{sbred}53.65 & \color{sbred}42.23 \\
            \midrule
            \resnethundred & \imagenetk{}    &   1200000 & 77.21 & 27.58 & 4.47 & 39.81 & 65.56  & 36.63  \\
            \bottomrule
        \end{tabular}
    }
\end{table}

\section{On the choice of \imagenet{}}\label{app:choice_imagenet}
We choose zero-shot classification on \imagenet{} and its distribution shifts as the main object of study four our work. Our analysis is agnostic to these choices and one could potentially use other datasets like iWILDCam 2021~\citep{beery2021iwildcam}, FMoW~\citep{christie2018functional}, MS-COCO~\citep{lin2015microsoft} and Flickr30k~\citep{yonglong_tian_divide_2021} and on tasks like image retrieval. One reason why we do not investigate this in our paper for the reason that several ImageNet distribution shifts are more human aligned than for the aforementioned datasets. For instance, there's a perceptual demarcation between a sketch of a dog (\imagenets{}) and a natural image of a dog (\imagenett{}). In contrast, it is unclear what the distribution shift of Flickr-30k is from MS-COCO. Additionally, retrieval tasks are more complex and highly sensitive to captions, demanding an analysis that factors in both images and texts. Another reason to choose zero-shot classification and \imagenet{} is that CLIP demonstrated unprecedented performance on ImageNet-based distribution shifts \citep{radford2021learning}. Moreover, we find iWILDCam and FMoW datasets problematic since CLIP’s (ViT-B/32) zero-shot performance on them is rather low (7.45\% and 12.96\%) \citep{ilharco_gabriel_2021_5143773}. We therefore chose \imagenet{} and its distribution shifts for our study and leave analysis on other datasets for future work.

\section{Similarity analysis on \celeba{} and \waterbirds{}}\label{app:other_datasets}
Secs.~\ref{sec:difficulty}~and~\ref{sec:experiments} only considered test sets with a clear distribution shift with respect to \imagenett{}. However, the general method outlined in Sec.~\ref{sec:gap} is dataset-agnostic. To illustrate this point, we here consider test sets that exhibit distribution shifts with respect to other datasets.
Specifically, we consider \celeba{}~\citep{liu2015faceattributes} and \waterbirds{}~\cite{sagawa2019distributionally}.

\begin{figure}[h]
    \centering
    \includegraphics[width=\textwidth]{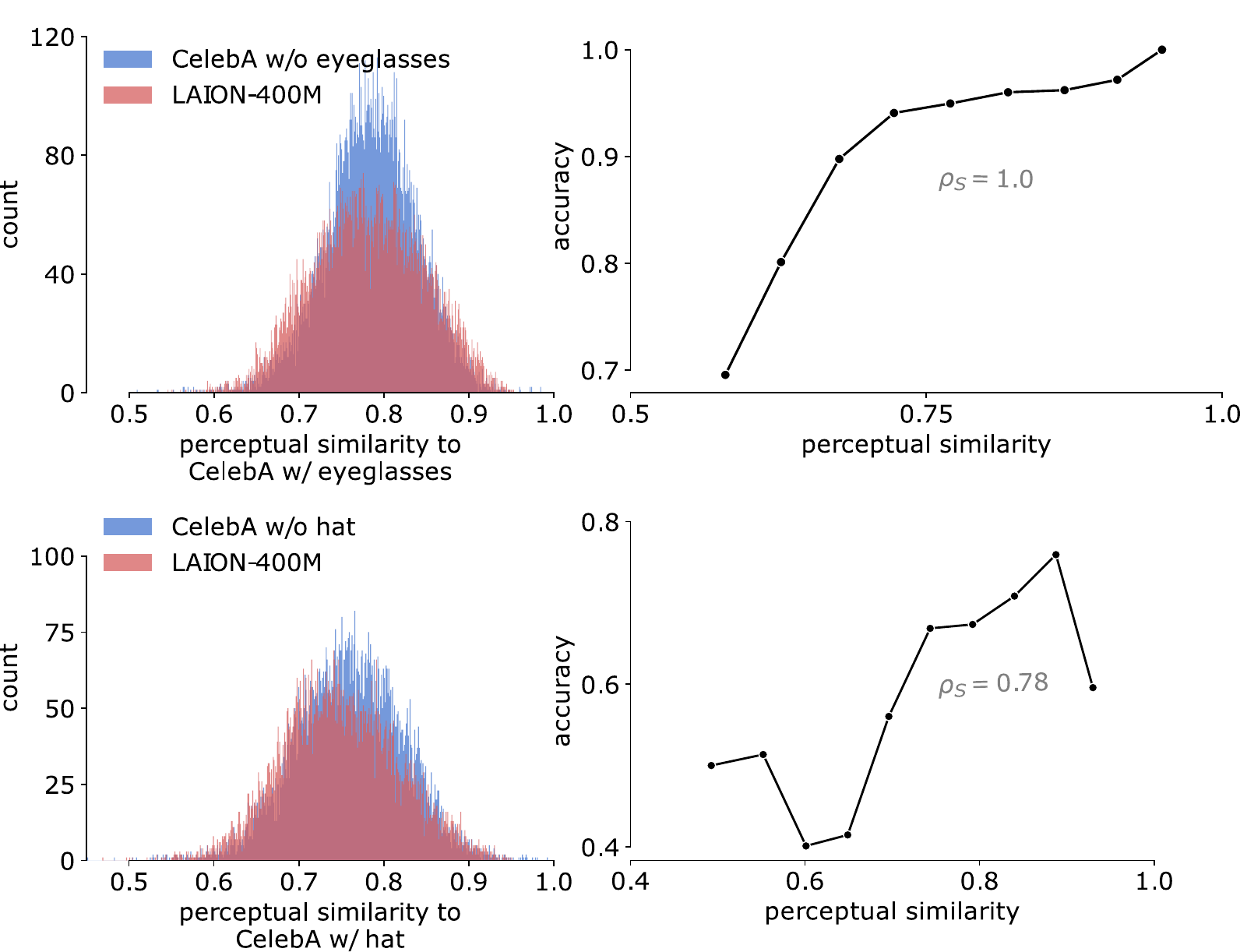}
    \caption{
        \textbf{nearest-neighbor similarity distribution and correlation to zero-shot accuracy for \celeba}.
        \textbf{Left}: The histogram shows the similarity of samples in \celeba{}-w/-eyeglasses to their nearest neighbors in \laionfour{} (red) and \celeba{}-w/o-eyeglasses (blue).
        \textbf{Right}: The strong correlation between perceptual similarity of test points to nearest neighbors in \laionfour{} samples and CLIP's top-1 classification accuracy on Male/Female classification indicates that differences in similarity can be expected to impact the performance of LAION-trained models. Data points in the correlation plot are averaged over bins (interval = 0.05) of the red histograms in the left plot.
    }\label{fig:sim_celeba}
\end{figure}

\paragraph{\celeba{}}
This dataset contains \SI{202599}{} celebrity images with 40 annotated attributes~\citep{liu2015faceattributes}. \citet{gannamaneni2023investigating} showed that CLIP could zero-shot predict many attributes with high accuracy (between \SI{53}{\%} and \SI{97}{\%} top-1 accuracy). We can split the data along each attribute to obtain training and test sets with a specific distribution shift. In Fig.~\ref{fig:sim_celeba}, we repeat the similarity analysis from Sec.~\ref{sec:nn_sim_distribution} for \celeba{} splits along the `eyeglasses' and `hat' attributes. The distribution of nearest-neighbor similarities to the test set \celeba-w/-eyeglasses (\celeba-w/-hat) differs between \laionfour{} and \celeba{}-w/o-eyeglasses (\celeba-w/o-hat). We again observe a strong correlation between the similarity of a test sample to \laionfour{} and CLIP's zero-shot accuracy in predicting the `gender' attribute.

\begin{figure}[h]
    \centering
    \includegraphics[width=\textwidth]{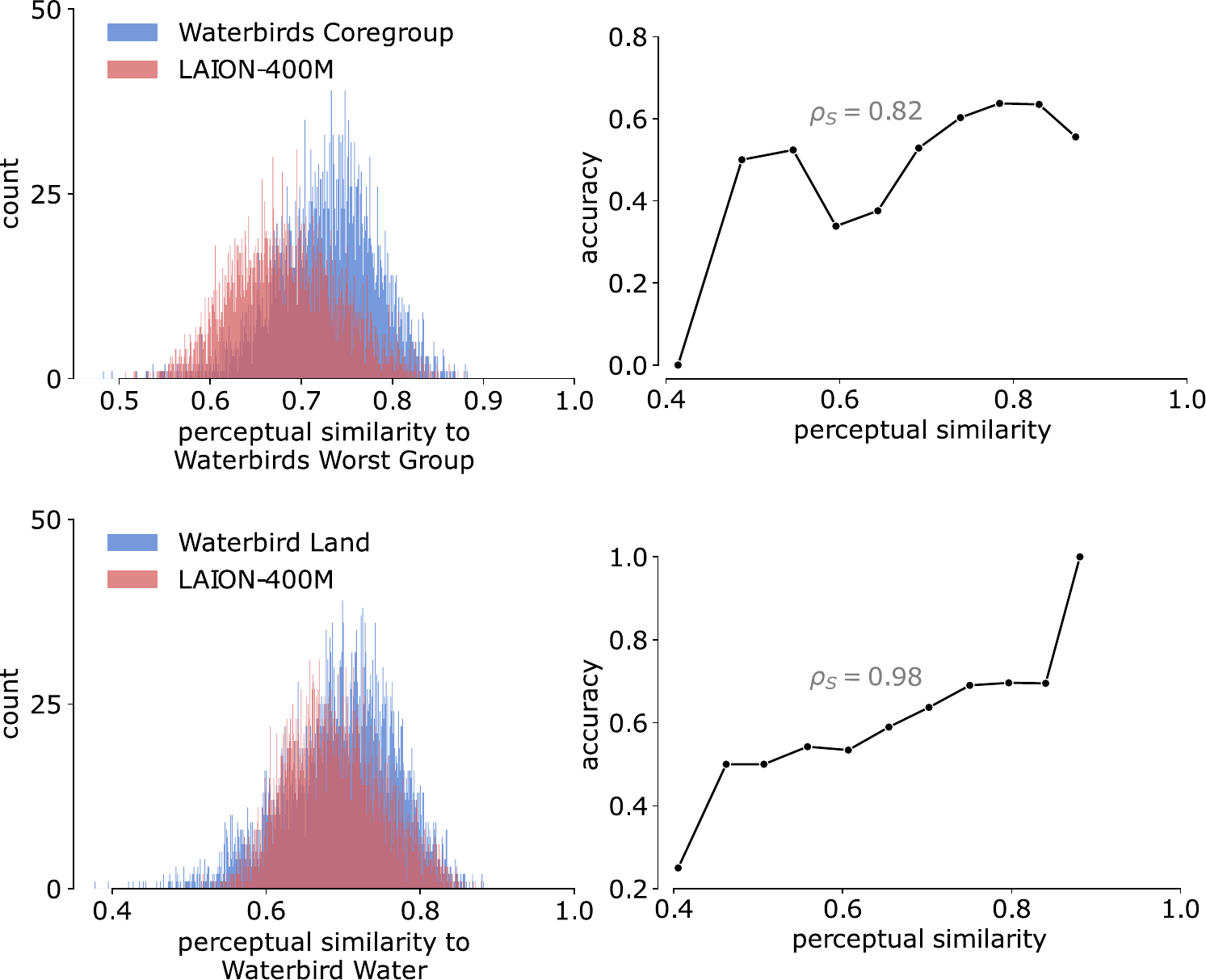}
    \caption{
        \textbf{nearest-neighbor similarity distribution and correlation to zero-shot accuracy for \waterbirds{}}.
        \textbf{Left}: The histogram shows the similarity of samples in \waterbirds{}-water to their nearest neighbors in \laionfour{} (red) and \waterbirds{}-land (blue).
        \textbf{Right}: The strong correlation between perceptual similarity of test points to nearest neighbors in \laionfour{} samples and CLIP's top-1 classification accuracy on landbird/waterbird classification indicates that differences in similarity can be expected to impact the performance of LAION-trained models. Data points in the correlation plot are averaged over bins (interval = 0.05) of the red histograms in the left plot.
    }\label{fig:sim_waterbird}
\end{figure}

\paragraph{\waterbirds{}}
We generate two sets of splits of this dataset.
We first split by background (land or water) and obtain a distribution shift from Waterbirds-land with 7051 images (6220 landbirds, 831 waterbirds) to Waterbirds-water with 4737 images (2905 landbirds, 1832 waterbirds).
We then split into the core group and the worst group. The core group consists of 8052 images of landbirds on land or waterbirds on water. The worst group consists of 3736 images of landbirds on water or waterbirds on land. In Fig.~\ref{fig:sim_waterbird}, we repeat the similarity analysis from Sec.~\ref{sec:nn_sim_distribution} for \waterbirds{} splits along the land/water and core/worst-group distribution shifts. The distribution of nearest-neighbor similarities to the test set \waterbirds{}-water (\waterbirds{}-worst group) differs between \laionfour{} and \waterbirds{}-land (\waterbirds{}-core group), and we again observe a strong correlation between the similarity of a test sample to \laionfour{} and CLIP's zero-shot accuracy in predicting the landbird/waterbird class.

\newpage 
\section{Comparing embedding metrics}\label{app:comparing_metrics}
To our knowledge, \emph{perceptual similarity} as measured in \vitbsixteenp{}'s image embedding space is a leading metric to capture the semantic and stylistic similarity between images. While we found this metric to align well with our intuitive notion of similarity and believe it to have captured the vast majority of highly similar images (see also Appx.~\ref{sec:appendix-nn_vis_after_pruning} where we visualize the pruned datasets), we cannot guarantee that all highly similar images were removed.
While we believe, based on prior work \citep{dreamsim,abbas2023semdedup,datacomp,tipadapter} and our analysis, that CLIP's embedding space sufficiently captures relevant features, in this section, we ablate the influence of the embeddings used to compute the perceptual similarity.
Specifically, we compare \vitbsixteenp{} embeddings used throughout the main paper to the embeddings of ViT-L-14 trained on \laionfour{}, a much larger model.

\begin{figure}[h]
    \centering
    \includegraphics[width=\textwidth]{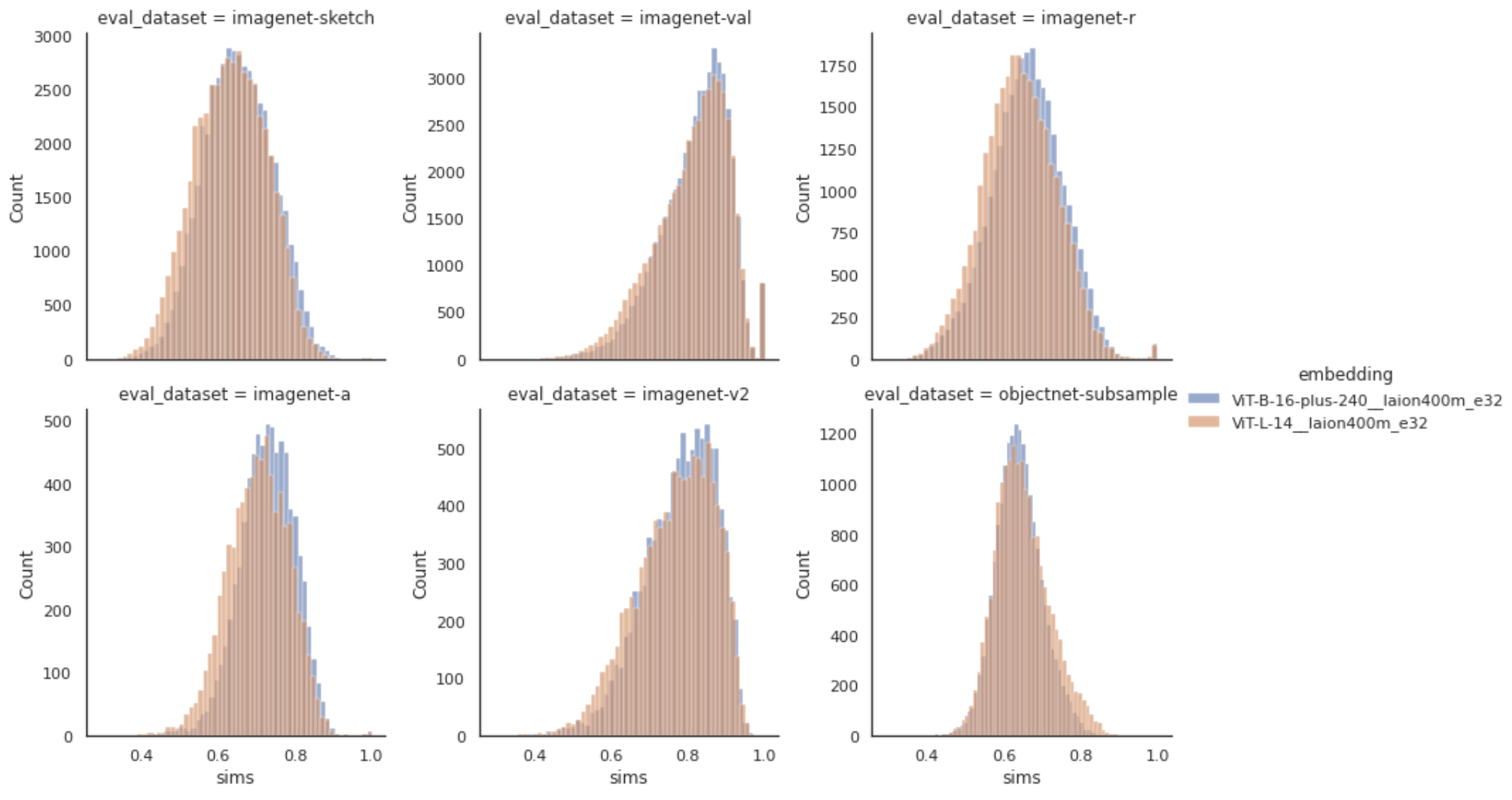}
    \caption{
        \textbf{Choice of embeddings has little impact on nearest-neighbor similarity distribution}.
        We compute nearest-neighbor similarities between \imagenett{} the six tests on embeddings produced by \vitbsixteenp{} and ViT-L-14 and find their distributions visually very similar across the bench.
    }\label{fig:sim_corr}
 \end{figure}

We compute the nearest-neighbor similarities of \imagenett{} to the test sets using either embeddings and compute their correlation.
We summarize the results in table~\ref{tab:correlation} and find that the correlation is strong across test sets.
Figure~\ref{fig:sim_corr} also shows histograms of nearest-neighbor similarities using either embedding, revealing that the similarity distributions are very similar.
Given these two comparisons, and considering that \vitbsixteenp{} embeddings are faster and cheaper to compute and have been shown to capture perceptual image similarities reasonably well by previous work~\citep{abbas2023semdedup}, we use them throughout our work. 

\begin{table}[h!]
    \centering
    \caption{
        \textbf{Choice of embeddings has little impact on nearest-neighbor similarities}.
        We compute nearest-neighbor similarities between \imagenett{} the six tests on embeddings produced by \vitbsixteenp{} and ViT-L-14 and find a strong correlation across the bench.
    }\label{tab:correlation}
    \sisetup{
        tight-spacing=true,
        detect-family=true,
        detect-weight=true,
        mode=text
    }
    \setlength{\tabcolsep}{0.4em}
    \renewcommand{\bfseries}{\fontseries{b}\selectfont} 
        \newrobustcmd{\B}{\bfseries}   
    \begin{tabular}{l S[table-format=1.2]}
        \toprule
        \B Dataset & $\rho_S$ \\
        \midrule
        ImageNet-Val  &  0.93 \\
        ImageNet-Sketch  & 0.87 \\
        ImageNet-R  &  0.90\\
        ImageNet-A  &  0.86\\
        ImageNet-V2  &  0.93\\
        ObjectNet  &  0.80\\
        \bottomrule
    \end{tabular}
    \vspace{10pt}
\end{table}

\section{Distribution of similarities of \laiontwo{} and \imagenett{} after pruning}\label{sec:after_pruning}
We analyzed nearest-neighbor similarity distribution of the test sets to \laiontwo{} and \imagenett{} (see Figs.~\ref{fig:histograms}~and~\ref{fig:histograms_app}). But what about the nearest-neighbor similarity distributions of \laiontwo{} and \imagenett{} to the test sets, especially after pruning? We now answer this question to understand better where the training points are situated with respect to the pruning boundary.

For each data point in the pruned dataset, we compute the maximum similarity to the respective test set and divide it by the test point's similarity gap (\ie, nearest-neighbor similarity of the test point to \laiontwo{} before pruning). We call this quantity \textit{normalized similarity}. Note that the normalized similarity values for samples in the pruned datasets are strictly smaller than 1.0 because samples with values greater than 1.0 are the ones that lie in the similarity gap and are pruned away.

\begin{figure}[h]
    \centering
    \includegraphics[width=1.15\textwidth]{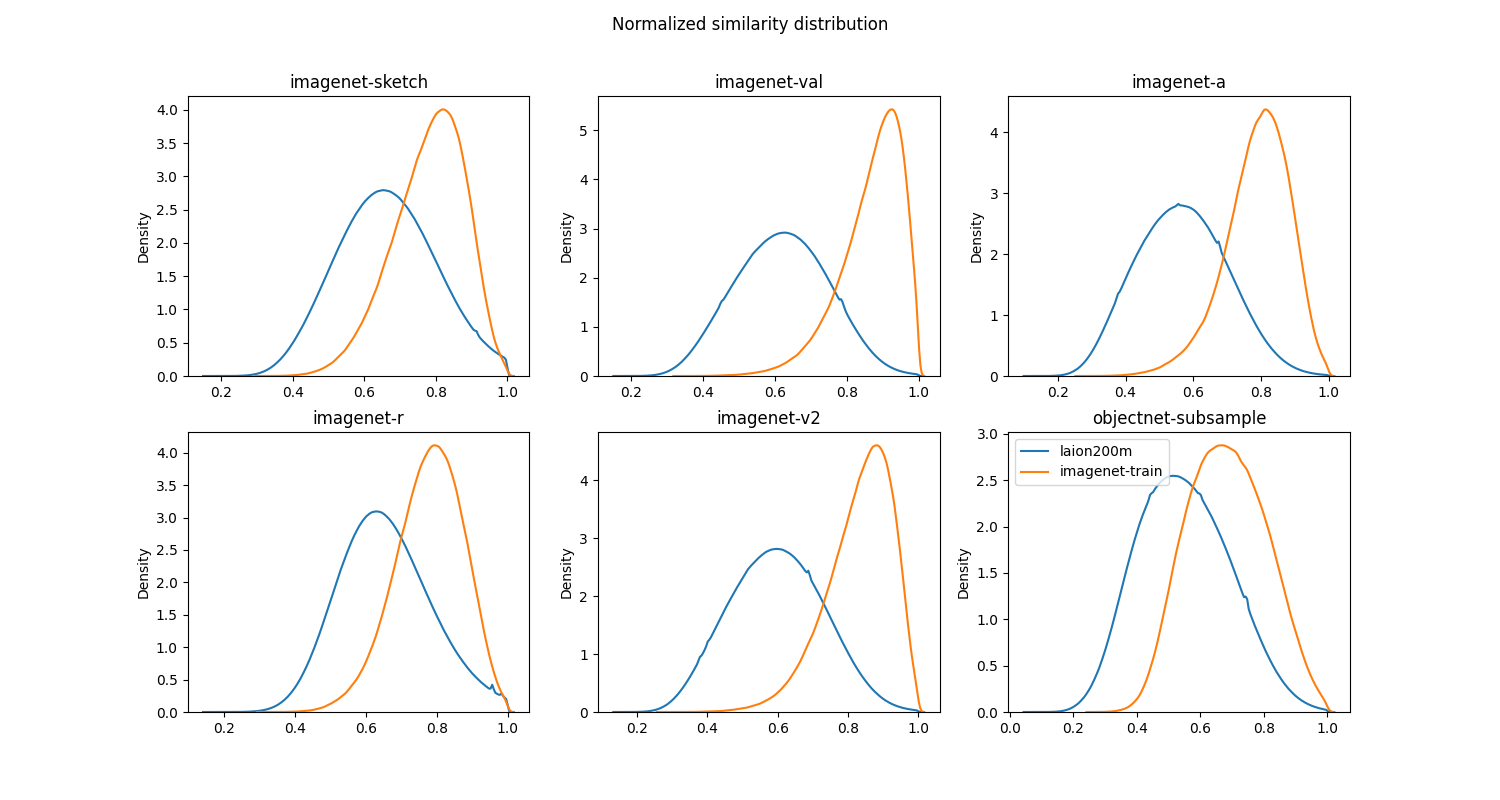}
    \caption{
        \textbf{Density of normalized similarity for LAION-pruned and \imagenett{}}.
        We observe a wider density function for LAION-pruned with a smaller mode. This indicates that \imagenett{} samples are generally more concentrated around the similarity gap.
    }\label{fig:norm_sim}
 \end{figure}

Plotting the density of normalized similarities in Fig.~\ref{fig:norm_sim} reveals that LAION-pruned has a much wider distribution with a smaller mode. Since normalized similarity closer to 1.0 indicates that the point lies closer to the similarity gap, a larger proportion of \imagenett{} samples are close to the similarity gap compared to the proportion observed in \laiontwo{}.

We also compare the total number of points that are close to the similarity gap (normalized similarity > 0.9) in Tab.~\ref{tab:boundary_total}. Due to LAION's scale, each pruned LAION split contains 5-20 times more data points close to the similarity gap than \imagenett{}. We expect that this large and diverse set of samples close to the boundary greatly dictates CLIP's performance.

\begin{table}[h]
    \centering
    \caption{
        \textbf{Total number of samples in LAION-pruned splits and ImageNet that lie near the boundary of the similarity gap for each test set}.
        Closeness is defined by a normalized similarity > 0.9.
        While LAION-pruned samples are less concentrated around the gap (see Fig.~\ref{fig:norm_sim}), LAION-pruned still has 5 to 20 times more samples close to the boundary than \imagenett{}.
    }\label{tab:boundary_total}
    \sisetup{
        tight-spacing=true,
        detect-family=true,
        detect-weight=true,
        mode=text
    }
    \setlength{\tabcolsep}{0.4em}
    \renewcommand{\bfseries}{\fontseries{b}\selectfont} 
    \newrobustcmd{\B}{\bfseries}   
    \begin{tabular}{l S[table-format=7] S[table-format=6]}
        \toprule
        \B Dataset & {\B \laiontwo{}} & {\B \imagenett{}} \\
        \midrule
        ImageNet-Sketch & 8859133 & 131087 \\ 
        ImageNet-Val & 2344086 & 531982 \\ 
        ImageNet-A & 1118150 & 138975 \\ 
        ImageNet-R & 7376362 & 121160 \\ 
        ImageNet-V2 & 1919398 & 326517 \\ 
        ObjectNet & 1558301 & 52277 \\ 
        \bottomrule
    \end{tabular}
    \vspace{10pt}
\end{table}

\section{Nearest neighbor visualizations}\label{sec:appendix-nn_vis}
 To generate nearest neighbors, we compute the nearest images of LAION in CLIP's (\vitbsixteenp{}) image embedding space for each test image. After removing duplicates and near-duplicates within LAION, we visualize the top six images.

\subsection{\laionfour{} vs \imagenett{}}\label{sec:appendix-nn_vis_before_pruning}
\begin{figure}[h]
    \centering
     \includegraphics[width=\textwidth]{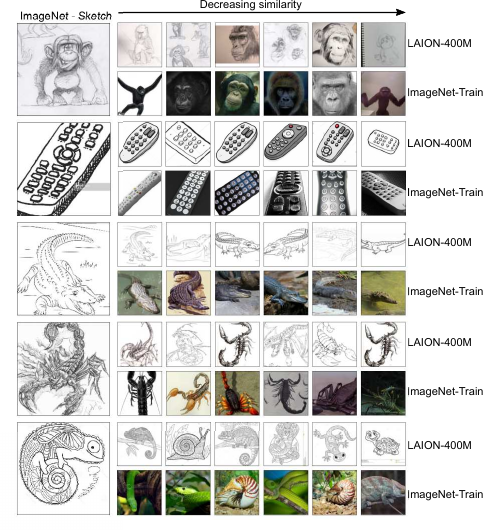}
     \caption{Nearest neighbors of \textit{randomly} sampled ImageNet-Sketch queries in \laionfour{} and \imagenett{} ordered by decreasing perceptual similarity. We omit duplicates within the nearest neighbors. Perceptual similarity is computed in CLIP's image embedding space and can be considered to measure the ``perceptual closeness'' of images in terms of content and style.}
     \label{fig:app_sketch_1}
 \end{figure}

\begin{figure}[h]
    \centering
     \includegraphics[width=\textwidth]{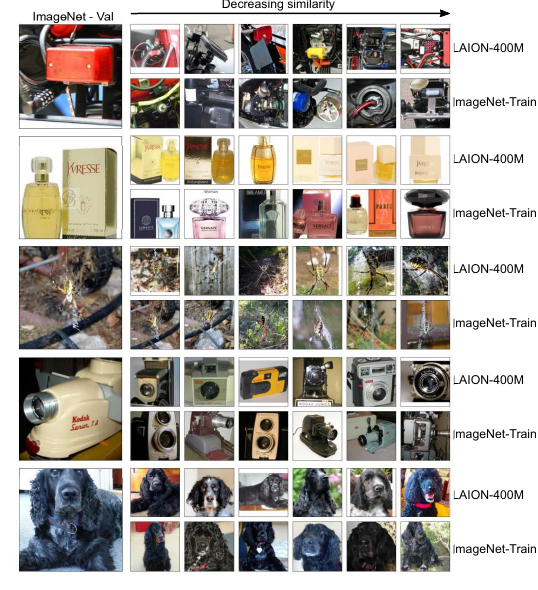}
     \caption{Nearest neighbors of \textit{randomly} sampled ImageNet-Val queries in \laionfour{} and \imagenett{} ordered by decreasing perceptual similarity. We omit duplicates within the nearest neighbors. Perceptual similarity is computed in CLIP's image embedding space and can be considered to measure the ``perceptual closeness'' of images in terms of content and style.}
     \label{fig:app_val_1}
 \end{figure}

\begin{figure}[h]
    \centering
     \includegraphics[width=\textwidth]{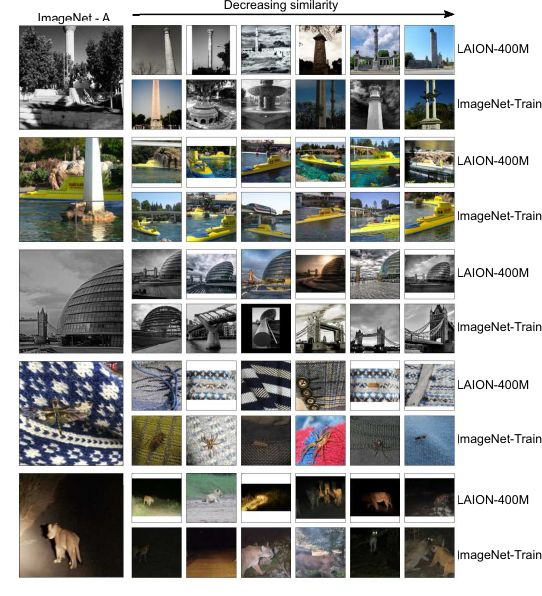}
     \caption{Nearest neighbors of \textit{randomly} sampled ImageNet-A queries in \laionfour{} and \imagenett{} ordered by decreasing perceptual similarity. We omit duplicates within the nearest neighbors. Perceptual similarity is computed in CLIP's image embedding space and can be considered to measure the ``perceptual closeness'' of images in terms of content and style.}
     \label{fig:app_imagenet_a}
 \end{figure}

\begin{figure}[h]
    \centering
     \includegraphics[width=\textwidth]{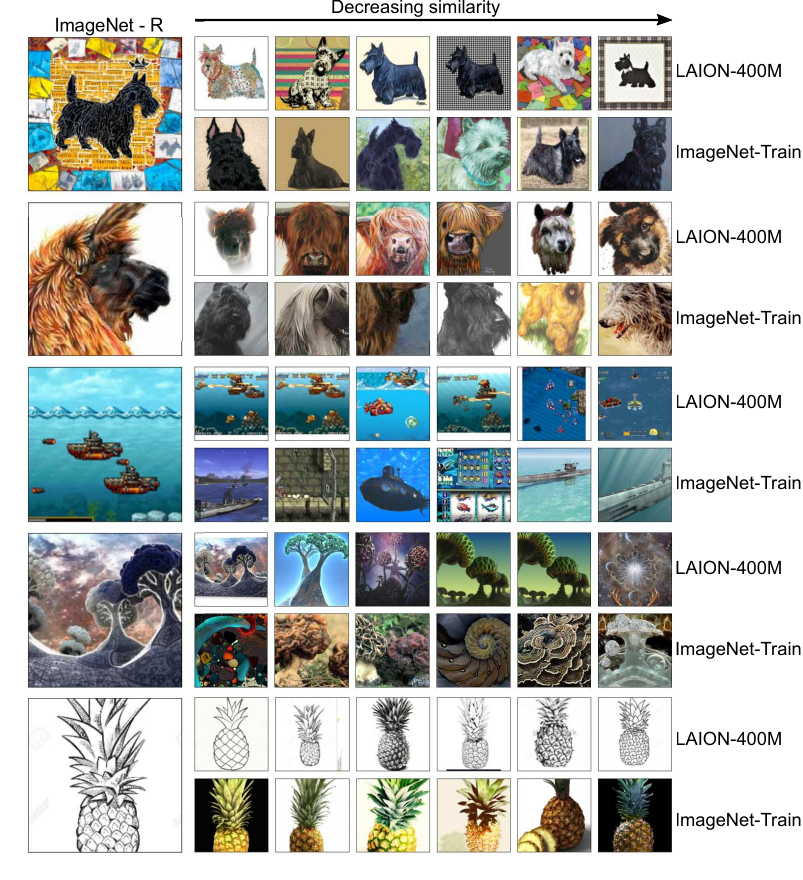}
     \caption{Nearest neighbors of \textit{randomly} sampled ImageNet-R queries in \laionfour{} and \imagenett{} ordered by decreasing perceptual similarity. We omit duplicates within the nearest neighbors. Perceptual similarity is computed in CLIP's image embedding space and can be considered to measure the ``perceptual closeness'' of images in terms of content and style.}
     \label{fig:app_imagenet_r}
 \end{figure}

\begin{figure}[h]
    \centering
     \includegraphics[width=\textwidth]{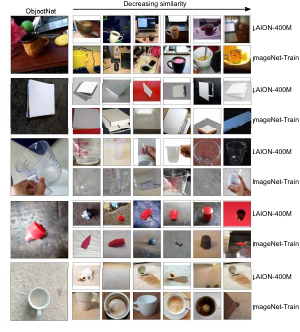}
     \caption{Nearest neighbors of \textit{randomly} sampled ObjectNet queries in \laionfour{} and \imagenett{} ordered by decreasing perceptual similarity. We omit duplicates within the nearest neighbors. Perceptual similarity is computed in CLIP's image embedding space and can be considered to measure the ``perceptual closeness'' of images in terms of content and style.}
     \label{fig:app_imagenet_objectnet}
 \end{figure}

\begin{figure}[h]
    \centering
     \includegraphics[width=\textwidth]{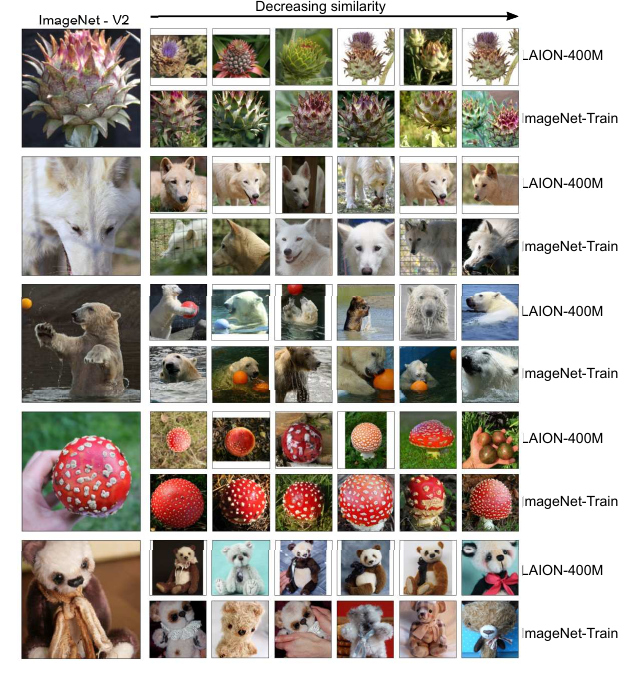}
     \caption{Nearest neighbors of \textit{randomly} sampled ImageNet-V2 queries in \laionfour{} and \imagenett{} ordered by decreasing perceptual similarity. We omit duplicates within the nearest neighbors. Perceptual similarity is computed in CLIP's image embedding space and can be considered to measure the ``perceptual closeness'' of images in terms of content and style.}
     \label{fig:app_imagenet_v2}
 \end{figure}

Just like in Fig.~\ref{fig:motivation}, we plot the nearest neighbors in \laionfour{} and \imagenett{} of random query images for each of the six datasets in Figs.~\ref{fig:app_sketch_1},~\ref{fig:app_val_1},~\ref{fig:app_imagenet_a},~\ref{fig:app_imagenet_r},~\ref{fig:app_imagenet_objectnet}, and~\ref{fig:app_imagenet_v2}.

\subsection{After pruning}\label{sec:appendix-nn_vis_after_pruning}
\begin{figure}[h]
    \centering
     \includegraphics[width=\textwidth]{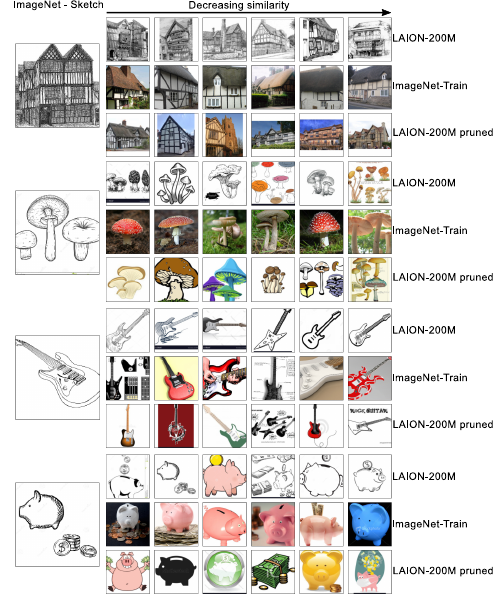}
     \caption{Nearest neighbors of \imagenets{} images in \laiontwo{}, \imagenett{}, and `sketch-pruned' (\laiontwo{} pruned) ordered by decreasing perceptual similarity. The query (base) images are \textit{randomly} sampled from the set of images that are more similar to \laiontwo{} than \imagenett{} to see the effect of pruning (see Tab.~\ref{tab:perc_laion_greater_in}). We omit duplicates within the nearest neighbors. Perceptual similarity is computed in CLIP's image embedding space and can be considered to measure the ``perceptual closeness'' of images in terms of content and style. \laiontwo{} clearly contains more similar images to samples in the test set compared to \imagenett{} or `sketch-pruned'.}
     \label{fig:app_after_sketch_1}
 \end{figure}

\begin{figure}[h]
    \centering
     \includegraphics[width=\textwidth]{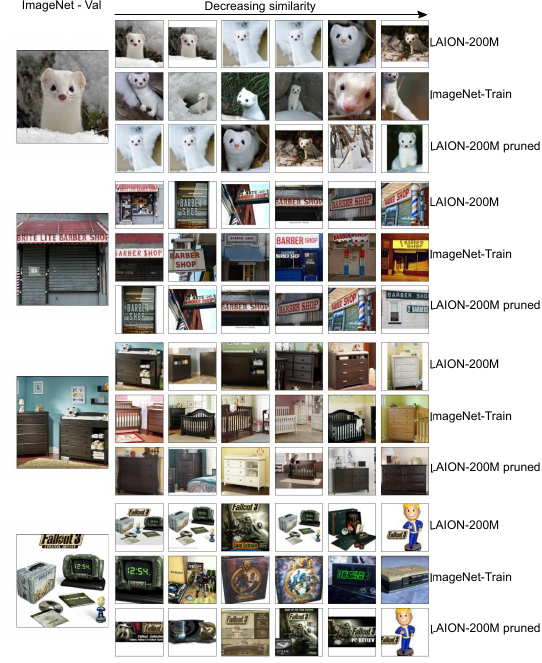}
     \caption{Nearest neighbors of \imagenetv{} images in \laiontwo{}, \imagenett{}, and `val-pruned' (\laiontwo{} pruned) ordered by decreasing perceptual similarity. The query (base) images are \textit{randomly} sampled from the set of images that are more similar to \laiontwo{} than \imagenett{} to see the effect of pruning (see Tab.~\ref{tab:perc_laion_greater_in}). We omit duplicates within the nearest neighbors. Perceptual similarity is computed in CLIP's image embedding space and can be considered to measure the ``perceptual closeness'' of images in terms of content and style. \laiontwo{} clearly contains more similar images to samples in the test set compared to `val-pruned'; \imagenett{} images are in-distribution to \imagenetv{} and, therefore, contain similar samples.}
     \label{fig:app_after_val_1}
 \end{figure}

\begin{figure}[h]
    \centering
     \includegraphics[width=\textwidth]{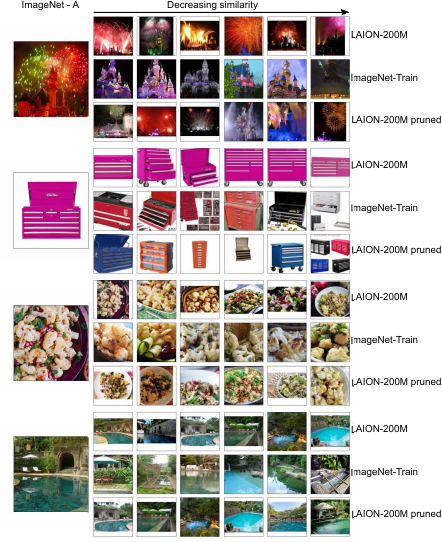}
     \caption{Nearest neighbors of ImageNet-A images in \laiontwo{}, \imagenett{}, and `a-pruned' (\laiontwo{} pruned) ordered by decreasing perceptual similarity. The query (base) images are \textit{randomly} sampled from the set of images that are more similar to \laiontwo{} than \imagenett{} to see the effect of pruning (see Tab.~\ref{tab:perc_laion_greater_in}). We omit duplicates within the nearest neighbors. Perceptual similarity is computed in CLIP's image embedding space and can be considered to measure the ``perceptual closeness'' of images in terms of content and style. \laiontwo{} clearly contains more similar images to samples in the test set compared to `val-pruned'; \imagenett{} images are in-distribution to \imagenetv{} and, therefore, contain similar samples.}
     \label{fig:app_after_imagenet_a}
 \end{figure}

\begin{figure}[h]
    \centering
     \includegraphics[width=\textwidth]{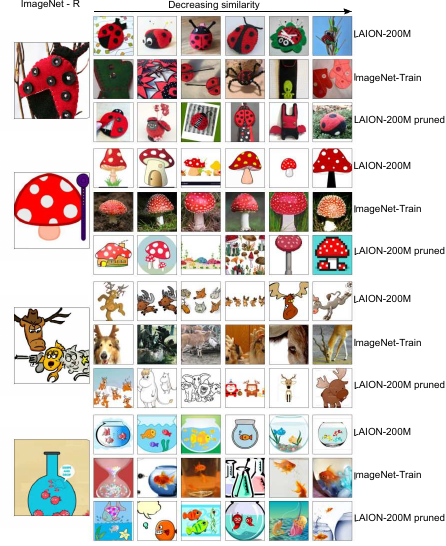}
     \caption{Nearest neighbors of ImageNet-R images in \laiontwo{}, \imagenett{}, and `r-pruned' (\laiontwo{} pruned) ordered by decreasing perceptual similarity. The query (base) images are \textit{randomly} sampled from the set of images that are more similar to \laiontwo{} than \imagenett{} to see the effect of pruning (see Tab.~\ref{tab:perc_laion_greater_in}). We omit duplicates within the nearest neighbors. Perceptual similarity is computed in CLIP's image embedding space and can be considered to measure the ``perceptual closeness'' of images in terms of content and style. \laiontwo{} clearly contains more similar images to samples in the test set compared to `val-pruned'; \imagenett{} images are in-distribution to \imagenetv{} and, therefore, contain similar samples.}
     \label{fig:app_after_imagenet_r}
 \end{figure}

\begin{figure}[h]
    \centering
     \includegraphics[width=\textwidth]{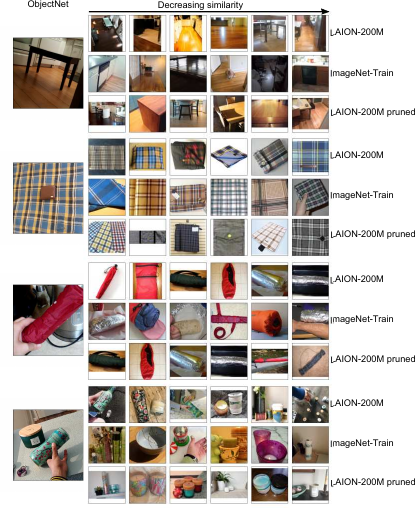}
     \caption{Nearest neighbors of ObjectNet images in \laiontwo{}, \imagenett{}, and `v2-pruned' (\laiontwo{} pruned) ordered by decreasing perceptual similarity. The query (base) images are \textit{randomly} sampled from the set of images that are more similar to \laiontwo{} than \imagenett{} to see the effect of pruning (see Tab.~\ref{tab:perc_laion_greater_in}). We omit duplicates within the nearest neighbors. Perceptual similarity is computed in CLIP's image embedding space and can be considered to measure the ``perceptual closeness'' of images in terms of content and style. \laiontwo{} clearly contains more similar images to samples in the test set compared to `val-pruned'; \imagenett{} images are in-distribution to \imagenetv{} and, therefore, contain similar samples.}
     \label{fig:app_after_imagenet_objectnet}
 \end{figure}

\begin{figure}[h]
    \centering
     \includegraphics[width=\textwidth]{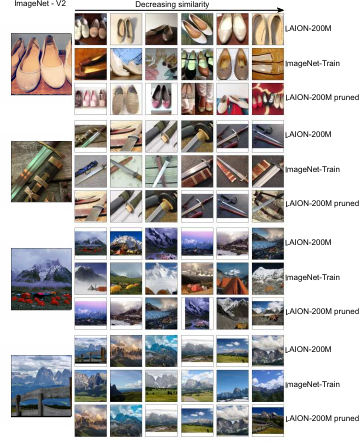}
     \caption{Nearest neighbors of ImageNet-V2 images in \laiontwo{}, \imagenett{}, and `objectnet-pruned' (\laiontwo{} pruned) ordered by decreasing perceptual similarity. The query (base) images are \textit{randomly} sampled from the set of images that are more similar to \laiontwo{} than \imagenett{} to see the effect of pruning (see Tab.~\ref{tab:perc_laion_greater_in}). We omit duplicates within the nearest neighbors. Perceptual similarity is computed in CLIP's image embedding space and can be considered to measure the ``perceptual closeness'' of images in terms of content and style. \laiontwo{} clearly contains more similar images to samples in the test set compared to `val-pruned'; \imagenett{} images are in-distribution to \imagenetv{} and, therefore, contain similar samples.}
     \label{fig:app_after_imagenet_v2}
 \end{figure}

Tab.~\ref{tab:perc_laion_greater_in} reports the percentage of images in each of the six datasets that have higher similarity to \laiontwo{}/\laionfour{} than \imagenett{}. For each of the six test sets, we randomly sample query images that are more similar to \laiontwo{} than \imagenett{} and plot the nearest neighbors in \imagenett{}, \laiontwo{}, and \laiontwo{} after pruning by the respective test in Figures~\ref{fig:app_after_sketch_1},~\ref{fig:app_after_val_1},~\ref{fig:app_after_imagenet_a},~\ref{fig:app_after_imagenet_r},~\ref{fig:app_after_imagenet_objectnet}, and~\ref{fig:app_after_imagenet_v2}.

\begin{table}[h]
    \centering
        \caption{\textbf{Percentage (\%) of points in the test datasets for which the nearest neighbor is in \laionfour{}/\laiontwo{} rather than \imagenett{}.}}\label{tab:perc_laion_greater_in}
    \sisetup{
        tight-spacing=true,
        detect-family=true,
        detect-weight=true,
        mode=text
    }
    \setlength{\tabcolsep}{0.4em}
    \renewcommand{\bfseries}{\fontseries{b}\selectfont} 
    \newrobustcmd{\B}{\bfseries}   
    \begin{tabular}{l S[table-format=5] S[table-format=2.2] S[table-format=2.2]}
        \toprule
        \B Dataset & {\B  Size} & {\B \laionfour{}} & {\B \laiontwo{}} \\
        \midrule
        ImageNet-Val  & 50000 & 16.80 & 14.88 \\
        ImageNet-Sketch  & 50889 & 97.94 & 97.45 \\
        ImageNet-R  & 30000 & 87.88 & 86.74 \\
        ImageNet-A  & 7500 & 47.39 & 45.53 \\
        ImageNet-V2  & 10000 & 38.95 & 35.48 \\
        ObjectNet  & 18574 & 63.24 & 61.62 \\
        \bottomrule
    \end{tabular}
    \vspace{10pt}
\end{table}



\end{document}